\documentclass{article}

\usepackage[accepted]{icml2022}
\usepackage[utf8]{inputenc} 
\usepackage[T1]{fontenc}    
\usepackage{hyperref}       
\usepackage{url}            
\usepackage{booktabs}       
\usepackage{amsfonts}       
\usepackage{nicefrac}       
\usepackage{microtype}      

\usepackage{microtype}
\usepackage{graphicx}
\usepackage{booktabs} 

\usepackage{natbib}
\usepackage{amssymb, amsmath,amsthm}
\usepackage{amsfonts}
\usepackage{algorithm}
\usepackage{algorithmic}
\usepackage{paralist}
\usepackage{multirow}
\usepackage{times}
\usepackage{wrapfig}

\usepackage{url,enumerate}
\usepackage{color,xcolor}
\usepackage{makeidx}  
\usepackage{amsmath,amssymb}
\usepackage{mathtools}
\usepackage[small, compact]{titlesec}
\usepackage{xspace}
\usepackage{epstopdf}
\usepackage{mathrsfs}
\usepackage{times}
\usepackage{enumerate}
\usepackage{color}
\usepackage{graphicx,epsfig}
\usepackage{amsmath,amssymb,xspace}
\usepackage{url}
\usepackage{bm}
\usepackage{bbm}
\usepackage{upgreek}
\usepackage{cleveref}
\usepackage{multirow}
\usepackage{ulem}
\usepackage{cancel}
\usepackage{subcaption}

\icmltitlerunning{\ours: A United Framework to Integrate Physics into Gaussian Processes}

\newcommand{\ours}{AutoIP\xspace}

\newcommand{\michael}[1]{}
\newcommand{\zhe}[1]{}
\begin{document}

\twocolumn[
\icmltitle{\ours: A United Framework to Integrate Physics into Gaussian Processes}



\icmlsetsymbol{equal}{*}

\begin{icmlauthorlist}
\icmlauthor{Da Long}{theU}
\icmlauthor{Zheng Wang}{theU}
\icmlauthor{Aditi S. Krishnapriyan}{UCB,Lawrence}
\icmlauthor{Robert M. Kirby}{theU}
\icmlauthor{Shandian Zhe}{theU}
\icmlauthor{Michael W. Mahoney}{UCB,Lawrence,ICSI}
\end{icmlauthorlist}

\icmlaffiliation{theU}{University of Utah}
\icmlaffiliation{UCB}{University of California, Berkeley}
\icmlaffiliation{Lawrence}{Lawrence Berkeley National Laboratory}
\icmlaffiliation{ICSI}{International Computer Science Institute}

\icmlcorrespondingauthor{Shandian Zhe}{zhe@cs.utah.edu}

\icmlkeywords{Machine Learning, ICML}

\vskip 0.3in
]



\printAffiliationsAndNotice{}  

\newcommand{\var}{{\rm var}}
\newcommand{\Tr}{^{\rm T}}
\newcommand{\vtrans}[2]{{#1}^{(#2)}}
\newcommand{\kron}{\otimes}
\newcommand{\schur}[2]{({#1} | {#2})}
\newcommand{\schurdet}[2]{\left| ({#1} | {#2}) \right|}
\newcommand{\had}{\circ}
\newcommand{\diag}{{\rm diag}}
\newcommand{\invdiag}{\diag^{-1}}
\newcommand{\rank}{{\rm rank}}
\newcommand{\nullsp}{{\rm null}}
\newcommand{\tr}{{\rm tr}}
\renewcommand{\vec}{{\rm vec}}
\newcommand{\vech}{{\rm vech}}
\renewcommand{\det}[1]{\left| #1 \right|}
\newcommand{\pdet}[1]{\left| #1 \right|_{+}}
\newcommand{\pinv}[1]{#1^{+}}
\newcommand{\erf}{{\rm erf}}
\newcommand{\hypergeom}[2]{{}_{#1}F_{#2}}
\newcommand{\tka}{{\tilde{\kappa} }}
\renewcommand{\a}{{\bf a}}
\renewcommand{\b}{{\bf b}}
\renewcommand{\c}{{\bf c}}
\renewcommand{\d}{{\rm d}}  
\newcommand{\e}{{\bf e}}
\newcommand{\f}{{\bf f}}
\newcommand{\g}{{\bf g}}
\newcommand{\h}{{\bf h}}
\renewcommand{\k}{{\bf k}}
\newcommand{\m}{{\bf m}}
\newcommand{\mb}{{\bf m}}
\newcommand{\n}{{\bf n}}
\renewcommand{\o}{{\bf o}}
\newcommand{\p}{{\bf p}}
\newcommand{\q}{{\bf q}}
\renewcommand{\r}{{\bf r}}
\newcommand{\s}{{\bf s}}
\renewcommand{\t}{{\bf t}}
\renewcommand{\u}{{\bf u}}
\renewcommand{\v}{{\bf v}}
\newcommand{\w}{{\bf w}}
\newcommand{\x}{{\bf x}}
\newcommand{\hx}{{\hat{\x}}}
\newcommand{\hf}{{\hat{\f}}}

\newcommand{\y}{{\bf y}}
\newcommand{\z}{{\bf z}}
\newcommand{\A}{{\bf A}}
\newcommand{\B}{{\bf B}}
\newcommand{\C}{{\bf C}}
\newcommand{\D}{{\bf D}}
\newcommand{\E}{{\bf E}}
\newcommand{\F}{{\bf F}}
\newcommand{\G}{{\bf G}}
\renewcommand{\H}{{\bf H}}
\newcommand{\I}{{\bf I}}
\newcommand{\J}{{\bf J}}
\newcommand{\K}{{\bf K}}
\newcommand{\hK}{\widehat{\K}}
\renewcommand{\L}{{\bf L}}
\newcommand{\M}{{\bf M}}
\newcommand{\N}{\mathcal{N}}  
\newcommand{\gp}{\mathcal{GP}}  
\newcommand{\Acal}{\mathcal{A}}
\newcommand{\Ocal}{\mathcal{O}}
\newcommand{\Dcal}{\mathcal{D}}
\newcommand{\Ycal}{\mathcal{Y}}
\newcommand{\Zcal}{\mathcal{Z}}
\newcommand{\Fcal}{\mathcal{F}}
\newcommand{\Vcal}{\mathcal{V}}
\newcommand{\Lcal}{\mathcal{L}}
\newcommand{\Tcal}{\mathcal{T}}
\newcommand{\Gcal}{\mathcal{G}}
\newcommand{\Hcal}{\mathcal{H}}
\newcommand{\Scal}{\mathcal{S}}

\renewcommand{\O}{{\bf O}}
\renewcommand{\P}{{\bf P}}
\newcommand{\Q}{{\bf Q}}
\newcommand{\R}{{\bf R}}
\renewcommand{\S}{{\bf S}}
\newcommand{\T}{{\bf T}}
\newcommand{\U}{{\bf U}}
\newcommand{\V}{{\bf V}}
\newcommand{\W}{{\bf W}}
\newcommand{\X}{{\bf X}}
\newcommand{\hX}{{\hat{\X}}}
\newcommand{\Y}{{\bf Y}}
\newcommand{\Z}{{\bf Z}}
\newcommand{\Mcal}{{\mathcal{M}}}
\newcommand{\Wcal}{{\mathcal{W}}}
\newcommand{\Ucal}{{\mathcal{U}}}
\newcommand{\zhat}{{\widehat{\z}}}
\newcommand{\xhat}{{\widehat{x}}}
\newcommand{\that}{{\widehat{t}}}

\newcommand{\bfLambda}{\boldsymbol{\Lambda}}

\newcommand{\bsigma}{\boldsymbol{\sigma}}
\newcommand{\balpha}{\boldsymbol{\alpha}}
\newcommand{\bpsi}{\boldsymbol{\psi}}
\newcommand{\bphi}{\boldsymbol{\phi}}
\newcommand{\boldeta}{\boldsymbol{\eta}}
\newcommand{\Beta}{\boldsymbol{\eta}}
\newcommand{\btau}{\boldsymbol{\tau}}
\newcommand{\bvarphi}{\boldsymbol{\varphi}}
\newcommand{\bzeta}{\boldsymbol{\zeta}}

\newcommand{\blambda}{\boldsymbol{\lambda}}
\newcommand{\bLambda}{\mathbf{\Lambda}}
\newcommand{\bOmega}{\mathbf{\Omega}}
\newcommand{\bomega}{\mathbf{\omega}}
\newcommand{\bPi}{\mathbf{\Pi}}

\newcommand{\cov}{\text{cov}}
\newcommand{\bpi}{\boldsymbol{\pi}}
\newcommand{\btheta}{\boldsymbol{\theta}}
\newcommand{\bxi}{\boldsymbol{\xi}}
\newcommand{\bSigma}{\boldsymbol{\Sigma}}

\newcommand{\bgamma}{\boldsymbol{\gamma}}
\newcommand{\bGamma}{\mathbf{\Gamma}}

\newcommand{\bmu}{\boldsymbol{\mu}}
\newcommand{\1}{{\bf 1}}
\newcommand{\0}{{\bf 0}}

\newcommand{\bs}{\backslash}
\newcommand{\ben}{\begin{enumerate}}
\newcommand{\een}{\end{enumerate}}

 \newcommand{\notS}{{\backslash S}}
 \newcommand{\nots}{{\backslash s}}
 \newcommand{\noti}{{\backslash i}}
 \newcommand{\notj}{{\backslash j}}
 \newcommand{\nott}{\backslash t}
 \newcommand{\notone}{{\backslash 1}}
 \newcommand{\nottp}{\backslash t+1}

\newcommand{\notk}{{^{\backslash k}}}
\newcommand{\notij}{{^{\backslash i,j}}}
\newcommand{\notg}{{^{\backslash g}}}
\newcommand{\wnoti}{{_{\w}^{\backslash i}}}
\newcommand{\wnotg}{{_{\w}^{\backslash g}}}
\newcommand{\vnotij}{{_{\v}^{\backslash i,j}}}
\newcommand{\vnotg}{{_{\v}^{\backslash g}}}
\newcommand{\half}{\frac{1}{2}}
\newcommand{\msgb}{m_{t \leftarrow t+1}}
\newcommand{\msgf}{m_{t \rightarrow t+1}}
\newcommand{\msgfp}{m_{t-1 \rightarrow t}}

\newcommand{\proj}[1]{{\rm proj}\negmedspace\left[#1\right]}
\newcommand{\argmin}{\operatornamewithlimits{argmin}}
\newcommand{\argmax}{\operatornamewithlimits{argmax}}

\newcommand{\dif}{\mathrm{d}}
\newcommand{\abs}[1]{\lvert#1\rvert}
\newcommand{\norm}[1]{\lVert#1\rVert}
\newcommand{\hu}{\widehat{\u}}
\newcommand{\ie}{{\textit{i.e.,}}\xspace}
\newcommand{\eg}{{\textit{e.g.,}}\xspace}
\newcommand{\etc}{{\textit{etc.}}\xspace}
\newcommand{\EE}{\mathbb{E}}
\newcommand{\dr}[1]{\nabla #1}
\newcommand{\VV}{\mathbb{V}}
\newcommand{\sbr}[1]{\left[#1\right]}
\newcommand{\rbr}[1]{\left(#1\right)}
\newcommand{\zhec}[1]{{\textcolor{blue}{#1}}}
\newcommand{\cmt}[1]{}
\newcommand{\tg}{\widetilde{g}}
\newcommand{\tZ}{\widetilde{\Z}}
\newcommand{\teps}{\widetilde{\epsilon}}

\newcommand{\bi}{{\bf i}}
\newcommand{\bj}{{\bf j}}
\newcommand{\bK}{{\bf K}}

\begin{abstract}
Physical modeling is critical for many modern science and engineering applications. 
From a data science or machine learning perspective, where more domain-agnostic, data-driven models are pervasive, physical knowledge --- often expressed as differential equations --- is valuable in that it is complementary to data, and it can potentially help overcome issues such as data sparsity, noise, and inaccuracy. 
In this work, we propose a simple, yet powerful and general framework --- \ours, for Automatically Incorporating Physics --- that can integrate all kinds of differential equations into Gaussian Processes (GPs) to enhance prediction accuracy and uncertainty quantification. 
These equations can be linear or nonlinear, spatial, temporal, or spatio-temporal, complete or incomplete with unknown source terms, and so on. 
Based on kernel differentiation, we construct a GP prior to sample the values of the target function, equation-related derivatives, and latent source functions, which are all jointly from a multivariate Gaussian distribution. 
The sampled values are fed to two likelihoods: one to fit the observations, and the other to conform to the equation. 
We use the whitening method to evade the strong dependency between the sampled function values and kernel parameters, and we develop a stochastic variational learning algorithm. 
\ours shows improvement upon vanilla GPs in both simulation and several real-world applications, even using rough, incomplete equations.

\end{abstract}





\vspace{-2mm}
\section{Introduction}
\vspace{-2mm}

Physical modeling is omnipresent and critical to many modern science and engineering applications, including weather and climate forecasting, bridge design, etc. 
To model a physical system, one usually writes down a set of ordinary differential equations (ODEs) or partial differential equations (PDEs) that characterize the system behavior according to physical laws. 
One then identifies boundary and/or initial conditions and solves the equations, typically via numerical methods, to obtain the solution function on the domain of interest. 
The solution is then used in the subsequent steps, such as system evolution and design optimization.

Machine learning (ML) and data science use a different paradigm. 
Methods from these areas estimate or reconstruct target functions from observed data, rather than by solving physical equations. 
To learn target functions, ML methods typically optimize a data-dependent loss.
Nonetheless, one would hope that the knowledge reflected in physical models, especially in ODEs and PDEs, is valuable to ML, in that this knowledge characterizes the local behaviors or properties of the target function, which then extrapolate to the entire domain of interest. 
Hence, as a complementary information source,  physics knowledge can potentially help overcome data sparsity, noise, and inaccuracy in measurements of physical systems. 
These problems are ubiquitous in practice. 

One example of an effort along these lines is provided by so-called physics-informed neural networks (PINNs)~\citep{raissi2019physics}, which use neural networks (NNs) to try to solve physical differential equations. 
PINNs simultaneously fit the boundary/initial conditions and minimize a residual term to conform to the equation.  
That is, from an optimization perspective~\citep{Boyd04,NW06}, PINNs do \textit{not} solve a constrained optimization problem, where physical knowledge is included as a constraint.
Instead, they adopt a penalty method approach (as opposed to an augmented Lagrangian method), solving a (related but non-equivalent) soft-constrained problem.
Also, PINNs demand that the form of the equation be fully specified.
From the data science perspective, this might restrict their capability to leverage physics knowledge in a broader sense. 
That is, the knowledge within incomplete equations, \eg those including latent sources (functions), cannot be incorporated. 
In addition, the differentiation operators on the NN itself (in the residual) make the effective loss function quite complicated~\citep{krishnapriyan2021characterizing}, bringing challenges in optimization/training, robustness, and uncertainty quantification~\citep{EdwCACM22}. 
\michael{There are several related works, including the Chen paper.  Why do we position ourselves with respect to PINNs?  They don't.  They position themselves with respect to Cauchy and Laplace.  Can we position ourselves more generally?  Maybe we can position ourselves w.r.t. PINNs and Chen, moving some of the Chen text below up to here, showing how we improve/bridge both.}\zhe{Thanks, Michael. That is a great point. I wrote the paragraph just to introduce how the idea was developed and came out of the mind, from doing the works related to PINNs (through my experience and discussion with Aditi, etc.). But I do agree, a position under a broader foundation, like in scientific ML or physics, would be more exciting. I am thinking about how to introduce our work in this way.}

In this work, we consider incorporating physics knowledge into Gaussian processes (GPs).
GPs are a nonparametric Bayesian modeling approach, which is not only flexible enough to learn complex functions from data, but also is  convenient to quantify the uncertainty (due to their closed-form posterior distribution). 
In most cases, GPs perform well with simple kernels, \eg Square Exponential (SE),  without the need for complex architecture design and hyper-parameter tuning; and, in many cases, methods from randomized numerical linear algebra (RandNLA) ~\citep{Mah-mat-rev_JRNL,DM16_CACM,DM21_NoticesAMS} can be used to speed up traditional algorithms for GP training and computation.
\michael{Why do we mention the SE there?  And what does ``most'' mean?  That sounds like.  That sounds like we are severely limiting applicability.}\zhe{We list ``SE'' here just to give a concrete example. ``most'' means the practice of GP rarely needs complex, deep kernel designs.  SE is almost the by-default choice of GP works in the community.}To this end, we propose \ours, a framework for Automatically Incorporating Physics.
\ours can incorporate all kinds of differential equations into GPs to enhance their prediction accuracy and uncertainty estimates. 
These equations can be linear or nonlinear, spatial, temporal, or spatio-temporal, complete or incomplete, including unknown source terms and coefficients, and so on. 
\michael{Do we evaluate on all those types of equations and/or make explicit below what arises for each of those possibilities.}\zhe{Yes, we did. Nonlinear pendulum and Allen-Cahn are both nonlinear equations; one is temporal and the other is spatial-temporal. The equations for Motion capture (Sec. 4.3) and Meta pollution (Sec. 4.4) are both linear; We have also tested the cases with complete and incomplete equations.}In this way, we can boost GPs with various sorts of physics knowledge. 

In more detail, \ours first samples a set of collocation points in the domain to support the equation.
It then uses kernel differentiation to construct a GP prior.
This prior jointly samples from a multi-variate Gaussian distribution the values of the target function at the training inputs and the values of all the equation-related derivatives and latent sources at the collocation points. 
In this way, \ours couples the target function and its derivatives in a probabilistic framework, without the need for conducting differential operations on a nonlinear surrogate (like with NNs).  
Next, \ours feeds these samples to two likelihoods. 
One is to fit the training data. 
The other is a virtual Gaussian likelihood that encourages conformity to the equation. 
\michael{Why in the last two sentences is one ``to fit'' and the other encourages conformity''?  Is that informal or technical?  The asymmetry in the text suggests something that might not be right.}\zhe{This is just informal. Since equation is not the data, we want to distinct them in the text. We can definitely replace ``conform'' by ``fit''.}
Since any differential equation is a combination of derivatives and source functions (if needed), it is  straightforward to combine their sampled values correspondingly in the virtual likelihood. 
In doing so, we can flexibly incorporate any equation. 
For effective and efficient inference, \ours uses the whitening method to parameterize the latent random variables with a standard Gaussian noise, thereby evading their strong dependency on the kernel parameters. 
\ours then jointly estimates the kernel parameters and the posterior of the noise with a stochastic variational learning algorithm. We find an interesting insight that the approximate posterior process is still a GP but with a new kernel, which maintains the kernel differentiation property. 

We illustrate our \ours framework in several benchmark physical systems, including nonlinear pendulums and the Allen-Cahn equation. 
\michael{A bit more detail here on how we evaluate and why.} 
We tested it with both complete and incomplete equations. For the latter, we hid a part of the ground-truth equation and view it as a latent source. 
In both cases, our approach largely improves upon the standard GP ({\ie without  physics knowledge incorporated}) in  prediction accuracy and uncertainty estimate when doing extrapolation.  
Next, on two real-world benchmark datasets, Swiss Jura and CMU motion datasets, {we examined our approach when integrating a sensible physics model that includes latent sources. 
Here, the ground-truth governing equations are unknown. 
\ours shows better prediction accuracy and predictive log-likelihood, as compared with GP and latent force models, a classical approach that can integrate physics knowledge when Green's functions are available. }
\michael{Why are we comparing to vanilla GP?  That seems like a strawman, can we do better with a stronger baseline, both so we know how it performs, and to defend ourselves from reviewers.  (We are one month after submission, and we'll hear back in a month.  If there is a response-to-reviewer phase, we should have our answers ready, and we should be ready if they ask us for a stronger baseline, so let's do this now.)}\zhe{We also compared with LFM in real-world applications, another GP based model that can incorporate equations including latent forces. However, in the illustrative examples, LFM is not applicable, since LFM only applies to linear equations with analytical Green's functions. Yes, we indeed can consider adding more experiments and baselines.}

\vspace{-2mm}
\section{Gaussian Process Regression} 
\label{sec:background}
\vspace{-2mm}

Gaussian processes (GPs) are stochastic priors in function space. Due to their nonparametric nature, GPs can self-adapt to the complexity of the target function according to  data, \eg from simple multilinear to highly nonlinear, not restricted to a specific parametric form. 
Suppose we aim to learn a function $f:\mathbb{R}^d \rightarrow \mathbb{R}$. 
When we place a GP prior over $f(\cdot)$, it means that $f$ is sampled as a realization of a Gaussian process governed by some covariance function $\kappa(\cdot, \cdot)$, $f\sim \gp\left(m(\cdot), \kappa(\cdot, \cdot)\right)$ where $m$ is the mean function, often set as the constant zero. 
The covariance function captures the stochastic correlation between the function values in terms of their inputs, and it is often chosen as a kernel function. 
For example, a popular choice is the Square Exponential (SE) kernel with Automatic Relevance Determination (ARD), 
$\text{cov}\left(f(\x), f(\x')\right) = \kappa(\x, \x') = \exp\left(-\frac{1}{2}(\x - \x')^\top \diag(\frac{1}{\s})(\x-\x')\right)$, where $\s$ are the length-scales (kernel parameters). 
\michael{What is SE-ARD?  We haven't defined it, and it's probably not so well known.  Also, talking about it so early suggests our framework is pretty narrow in scope, working or being designed for it.} \zhe{SE-ARD is almost the standard choice of GP based papers in ML community. But one can freely use any other kernels, like Matern, in the same framework. Here we just want to give a concrete example. }
The finite projection of the GP is a collection of the values of $f(\cdot)$ at an arbitrary finite set of inputs.
This follows a multivariate Gaussian distribution, where the covariance matrix is the kernel matrix on the input~set. 

Consider a training dataset $\Dcal = (\X, \y)$, where  $\X=[\x_1, \cdots, \x_N]^\top$, $\y=[y_1, \cdots, y_N]^\top$, each $\x_n$ is an input, and $y_n$ is a noisy observation of $f(\x_n$). 
Then the function values at the training inputs, $\f = [f(\x_1), \cdots, f(\x_N)]^\top$, follow a multivariate Gaussian distribution, $p(\f|\X) = \N(\f|\0, \K)$ where each  $[\K]_{i,j}=\kappa(\x_i, \x_j)$. 
Given $\f$, we can use a noisy model to sample the observation $\y$.  
A Gaussian noise model is the commonly used  one, $p(\y|\f) = \N(\y|\f, \beta^{-1}\I)$ where $\beta$ is the inverse noise variance.  
We can then marginalize out $\f$ to obtain the marginal likelihood of $\y$, \ie evidence, 
\begin{align}
	p(\y|\X) = \N(\y|\0, \K + \beta^{-1}\I). \label{eq:gp}
\end{align} 
To learn the model, one often maximizes the evidence to estimate the kernel parameters and the  inverse noise variance $\beta$.   Given a new input $\x^*$, according to the GP prior, $[f(\x^*); \y]$ also follows a multivariate Gaussian distribution. Hence,  the posterior (or predictive) distribution of $f(\x^*)$ is a conditional Gaussian, 
\begin{align}
	p\big(f(\x^*)|\x^*, \X, \y\big) = \N\big(f(\x^*)|\mu^*, v^*\big),
	\label{eq:predictive}
\end{align}
where $\mu^* = \k_*^\top(\K + \beta^{-1}\I)^{-1}\y$, $v^*=\kappa(\x^*, \x^*) - \kappa_*^\top(\K + \beta^{-1}\I)^{-1}\k_*$ and  $\k_* = [\kappa(\x^*, \x_1), \cdots,\kappa(\x^*, \x_N)]^\top$.  
Due to the closed-form posterior, GP models are convenient for quantifying and reasoning about under uncertainty.

\vspace{-2mm}
\section{Our \ours Framework}
\vspace{-2mm}


\paragraph{Model.}

To boost GPs with physics knowledge, we propose \ours --- a framework for Automatically Incorporating Physics from all kinds of differential equations. 
Without loss of generality, we use a nonlinear, incomplete PDE in the Allen-Cahn family to illustrate the idea~\citep{allen1972ground}.\michael{Can we frame the presentation below more generally?  E.g., derive the equations and then show how it relates to the Allen-Cahn family.}
\michael{Also, we need a reference for the Allen-Cahn family.}
\michael{And I don't know why we are presenting the Allen-Cahn family and how general it is, so we need to explain that.  I don't think most ML people would be able to take the derivation below and apply it to another PDE, e.g., knowing what is peculiar to this and what holds for \ours more generally.}\zhe{Thanks for your suggestion, Michael. The major concern is that if we first introduce the general or abstract definition of operators without explicitly telling how it is done (say with a concrete example) , the reviewers might not be able to understand our methodology and then start to complain. I  got many such comments before. Some comments just said they don't understand and so the paper needs to rewrite. I will consider how to explain our model with a more general narrative.}This PDE takes the form 
\begin{align}
	\partial_t u  -  \nu \cdot \partial_x^2 u + \gamma\cdot u(u^2-1) + g(x, t) = 0, \label{eq:pde-1}
\end{align} 
where the target function $u(x,t)$ is a spatial-temporal function, $g(x, t)$ is an unknown source term, and $\nu$ and $\gamma$ are coefficients.  
Note that $u(u^2-1)$ is a nonlinear term.  
Suppose we are given $N$ training examples, $\Dcal = \{(\z_1, y_1), \ldots, (\z_N, y_n)\}$, where each $\z_n = (x_n, t_n)$. 
We want our learned function not only to fit the observations but also to conform to \eqref{eq:pde-1}, \ie to the known physics. 
To this end, we sample a set of $M$ collocation points $\widehat{\Zcal} = \{\zhat_1, \ldots, \zhat_M\}$ in the domain of interest (\eg $[0, 2\pi]\times [0, 1]$), and we augment the GP model to encourage the L.H.S of the equation \eqref{eq:pde-1} evaluated at $\widehat{\Zcal}$ to be close to zero. 

Specifically, we first construct a GP prior over $u$, $g$ and the equation-related derivatives, \ie $\partial_t u$ and $\partial_x^2 u$.  
Naturally, we can sample $u \sim \gp\left(0, \kappa_u(\cdot, \cdot)\right)$ and $g \sim \gp\left(0, \kappa_g(\cdot, \cdot)\right)$. 
The key observation is that once $u$ is drawn, all of its derivatives are determined --- we do not need to draw them from separate GPs. 
The covariance and cross-covariance among $u$ and its derivatives can be obtained from $\kappa_u$ via kernel differentiation~\citep{williams2006gaussian}, 
\begin{align}
	\cov\left(u(\z_1), u(\z_2)\right) &= \kappa_u(\z_1, \z_2), \notag \\
	\cov\left(\partial_t u(\z_1), \partial_t u(\z_2)\right) &= \frac{\partial^2 \kappa_u(\z_1, \z_2)}{\partial t_1\partial t_2}, \notag \\
	\cov\left(\partial_x^2 u(\z_1), \partial_x^2 u(\z_2)\right) &= \frac{\partial^4 \kappa_u(\z_1, \z_2)}{\partial x_1^2\partial x_2^2}, \notag \\
	\cov\left(\partial_t u(\z_1), \partial_x^2 u(\z_2)\right) &= \frac{\partial^3 \kappa_u(\z_1, \z_2)}{\partial t_1\partial x_2^2}, \notag \\
	\cov\left(\partial_t u(\z_1),  u(\z_2)\right) &=  \frac{\partial \kappa_u(\z_1, \z_2)}{\partial t_1}, \notag \\
	\cov\left(\partial_x^2 u(\z_1),  u(\z_2)\right) &=  \frac{\partial^2 \kappa_u(\z_1, \z_2)}{\partial x_1^2}, \label{eq:cov-list}
\end{align}
where $\z_1 = (x_1, t_1)$ and $\z_2 = (x_2, t_2)$ are two arbitrary inputs (we abuse notation a bit here for convenience  --- the points $\z_1$ and $\z_2$ are different from the training inputs $\Zcal = \{\z_1, \ldots, \z_N\}$.) 
In general, we can obtain the covariance of two arbitrary derivatives (of the same function) by taking the partial derivatives of the original kernel with respect to the corresponding inputs (using the same order). 
Since the commonly used kernels, \eg the SE kernel, are quite simple, we can obtain their derivatives analytically and directly apply the result for computation.\michael{How commonly-used is that in ML?}\zhe{It is usually used as the default choice. Most GP papers simply use the SE kernel to demonstrate the performance} We denote the values of the target function at the training inputs by $\u = \left(u(\z_1), \ldots, u(\z_N)\right)^\top$, the values of $u$ and $u$'s derivatives at the collocation points by $\hu = \left(u(\zhat_1), \ldots, u(\zhat_M)\right)^\top$, $\hu_t = \left(\partial_t u(\zhat_1), \ldots, \partial_t u(\zhat_M)\right)^\top$ and $\hu_{xx} = \left(\partial^2_x u(\zhat_1), \ldots, \partial^2_x u(\zhat_M)\right)^\top$, and the values of the latent source term at the collocation points by $\g = \left(g(\zhat_1), \ldots, g(\zhat_N)\right)^\top$. 
Then, we can leverage the covariance functions in \eqref{eq:cov-list} and $\kappa_g$ to construct a joint Gaussian prior over $\f = [\u; \hu; \hu_t; \hu_{xx}; \g]$,
\begin{align}
	p(\f) = \N(\f | \0, \bSigma). \label{eq:joint-gp}
\end{align} 
The covariance matrix $\bSigma$ is block-diagonal, including a dense block for $[\u; \hu; \hu_t;\hu_{xx}]$ computed from \eqref{eq:cov-list} and another dense block for $\g$ computed via $\kappa_g$. 
Note that we can further model the covariance between $u$ and $g$ if we have more prior knowledge. 
Here, we consider the general case that assumes they are sampled from two independent GPs. 

Next, we feed the sampled $\f$ to two data likelihoods. 
\michael{What is ``them'' in that sentence?  Do you mean ``it'', i.e., $\f$?}
One is to fit the actual observations from a Gaussian noise model, 
\begin{align}
	p(\y|\f) = \N(\y|\u, \beta^{-1}\I) \label{eq: lk1}.
\end{align}
The other is a virtual Gaussian likelihood that integrates the physics knowledge into the differential equation \eqref{eq:pde-1}, as
\begin{align}
	\hspace{-3mm}
	p(\0|\f) 
	\hspace{-1mm}
	= 
	\hspace{-1mm}
	\N(\0|\hu_t - \nu \hu_{xx} + \gamma \hu\circ(\hu\circ\hu-\1) 
	\hspace{-1mm}
	+ 
	\hspace{-1mm}
	\g, v\I), \label{eq:lk2}
\end{align}
where $v$ is the variance and $\circ$ is element-wise product. 
\michael{Make that and other such equations one line if we have a full one-column format, otherwise it looks funny.}\zhe{Thanks. done.}
As we can see, the mean of the Gaussian in \eqref{eq:lk2} is the evaluation of the L.H.S of \eqref{eq:pde-1} at the collocation points. The variance $v$ indicates how it is close to zero. The smaller $v$ is, the more consistent the sampled functions are with the differential equation. In practice, we can either tune $v$ or learn $v$ to enforce the conformity to a certain degree. Finally, the joint probability of our model is given by
\begin{align}
    &
    \hspace{-3mm}
    p(\f, \y, \0) = \N(\f|\0, \bSigma)\N(\y|\u, \beta^{-1}\I) \notag \\
    &
    \hspace{3mm}
    \cdot \N(\0|\hu_t - \nu \hu_{xx} + \gamma \hu\circ(\hu\circ\hu-\1) + \g, v\I).
\end{align}
As we can see, by leveraging the kernel differentiation, our model constructs a GP prior to sample jointly the target function and all the basic components of the differential equation, \ie all kinds of derivatives and latent source terms (if needed). 
We naturally couple them into a probabilistic framework, without the need for taking explicit differentiation over some (complex) function surrogates.  
\michael{There was passing reference to PINNs here, which I removed.  If it is important enough, we should have a passing reference to both the PINNs and the Chen paper.}\zhe{Thanks.}Then, through the virtual Gaussian likelihood \eqref{eq:lk2}, we can combine these components following arbitrary differential equation (in the mean) to encode the physics knowledge. 
If there are unknown coefficients, \eg $\nu$ and $\gamma$ in \eqref{eq:pde-1}, we can estimate them jointly during model inference. 
While simple, our augmented GP is flexible enough to incorporate a variety of differential equations to benefit learning and prediction.

\paragraph{Algorithm.}

In general, the exact posterior of the latent random variables $\f$ is intractable to compute or marginalize out (as in standard GP regression), because the virtual likelihood \eqref{eq:lk2} couples the components of $\f$ to reflect the equation, which can be nonlinear and nontrivial. 
Hence, we develop a general variational inference algorithm to estimate jointly the posterior of $\f$ and kernel parameters, inverse noise variance $\beta$, $v$, etc. 
However, we found that a straightforward implementation to optimize the variational posterior $q(f)$  often gets stuck at an inferior estimate. 
This might be due to the strong coupling of $\f$ and the kernel parameters in the prior~\eqref{eq:joint-gp}. 
To address this issue, we use the whitening method~\citep{murray2010slice} in MCMC sampling. That is, we parameterize $\f$ with a Gaussian noise, 
\begin{align} 
	\f = \A\boldeta \label{eq:white}
\end{align} 
where $\boldeta \sim \N(\0, \I)$, and $\A$ is the Cholesky decomposition of the covariance matrix $\bSigma$, \ie $\bSigma = \A\A^\top$. Therefore, the joint probability of the model can be rewritten as 
\begin{align}
	p(\text{Joint}) = \N(\boldeta|\0, \I)p(\y|\L\boldeta) p(\0|\L\boldeta).
\end{align}
See \eqref{eq: lk1} and \eqref{eq:lk2} for $p(\y|\L\boldeta)$ and $p(\0|\L\boldeta)$, respectively. We then introduce a Gaussian variational posterior for the noise, $q(\boldeta) = \N(\boldeta|\bmu, \L\L^\top)$, where $\L$ is a lower-triangular matrix to ensure the positive definiteness of the covariance matrix.  Since the prior of $\boldeta$ is the standard normal distribution, it does not depend on the kernel parameters any more. We then construct a variational evidence lower bound,
\begin{align}
	\Lcal &= -\text{KL}\left(q(\boldeta) \| \N(\boldeta|\0, \I)\right)  \notag \\
	& + \EE_q\left[\log p(\y|\L\boldeta)\right] + \EE_q\left[\log(p(\0|\L\boldeta))\right],
\end{align} 
where $\text{KL}(\cdot \|\cdot)$ is the Kullback-Leibler divergence.  
We maximize $\Lcal$ to estimate $q(\boldeta)$ and the other parameters. We  use the reparameterization trick~\citep{kingma2013auto} to conduct stochastic optimization.  Once we obtain $q(\boldeta)$, from \eqref{eq:white} we can immediately obtain the variational posterior of $\f$,  $q(\f) = \N(\f|\A \bmu, \A\L\L^\top\A^\top)$, according to which we can compute the predictive distribution of the function values at new inputs.  
We do not consider the computational challenge when the number of training examples ($N$) and/or collocation points ($M$) is big.
However, one can extend our algorithm to a variety of sparse GP frameworks, \eg~\citep{GPSVI13}, and/or use methods from RandNLA~\citep{Mah-mat-rev_JRNL,DM16_CACM,DM21_NoticesAMS} in order to handle large data.  

\zhe{some new insight after rebuttal, needs to polish later:}
\paragraph{Remarks.}
With the Gaussian variational approximation, the posterior process is still a GP and maintains the link between the function and its derivatives in terms of kernel differentiation. 
But the kernel has changed.  
This can be seen from the predictive distribution of an arbitrary finite set of the function and its derivative values, say $\h = (u(\z_1), u(\z_2), \partial_x u(\z_2), \ldots)$, which is, 
\begin{align}
&
\hspace{-1mm}
p(\h|\Dcal) \hspace{-0.5mm} = \hspace{-1.5mm} \int p(\h|\f) p(\f|\Dcal) \d \f \approx \hspace{-1.5mm} \int p(\h|\f) \N(\f|\m_f, \V_f)\d \f \notag \\
&= \hspace{-0.5mm} \N\left(\h | \m_h, \cov(\h, \h) \hspace{-0.5mm} - \hspace{-0.5mm} \cov(\h, \f)\cdot \B\cdot \cov(\f, \h)  \right), \hspace{-1mm} 
\label{eq:post}
\end{align}
where $\Dcal$ is the data (including $\y$ and the virtual observation $\0$), $\cov(\cdot)$ is obtained from the kernel $\kappa_u$ and its differentiation (see \eqref{eq:cov-list}), 
and $\m_f$ and $\V_f$ are the estimated posterior mean and covariance of $\f$, $\m_h = \cov(\h, \f) \bSigma^{-1} \m_f$, and $\B = \bSigma^{-1} - \bSigma^{-1}\V_f \bSigma^{-1} $.  

The result \eqref{eq:post} defines a GP for $u(\cdot)$ with a new kernel:  
$\overline{\cov}(u(\z_1), u(\z_2) ) = \rho(\z_1, \z_2) $, where $\z_1$ and $\z_2$ are two arbitrary inputs, $\rho(\z_1, \z_2) = \kappa_u(\z_1, \z_2) - \tka(\z_1, \Z)\cdot \B \cdot \tka(\z_2, \Z)$, $\Z = \{\Zcal, \widehat{\Zcal}\}$ are the corresponding inputs of $\f$, and $\tka(\z, \Z) = \cov(u(\z), \f)$, which applies   $\kappa_u$ or its partial derivatives over $\z$ and each input in $\Z$; see \eqref{eq:cov-list}. 
To verify if the link between $u$ and its derivatives is still there, we  examine the derivatives of the new kernel $k(\z_1, \z_2)$ w.r.t its inputs $\z_1$ and $\z_2$. Since {$\B$ and $\Z$ are both constant to the  inputs of $\rho$, the differentiation is only applied to $\kappa_u$ and $\tka$ on $\z_1$ and/or $\z_2$}.  
For example,  $\partial \rho(\z_1, \z_2)/\partial x_2 = \partial \kappa_u(\z_1, \z_2)/\partial x_2  -  \tka(\z_1, \Z) \cdot \B\cdot \partial \tka(\Z, \z_2)/\partial x_2 = \cov\left(u(\z_1), \partial_x u(\z_2)\right) - \cov(u(\z_1), \f) \cdot \B \cdot \cov(\f, \partial_x u(\z_2))$ (note $\z_2 = (x_2, t_2)$). 
Hence, the kernel differentiation gives the same covariance (between the function and its derivatives) as in the predictive distribution \eqref{eq:post}, \ie $\partial \rho(\z_1, \z_2)/\partial x_2  =  \overline{\cov}(u(\z_1), \partial_x u(\z_2))$.  
That means the kernel links are still maintained in the posterior/predictive process (\ie conditioned on the data and differential equation).


\vspace{-2mm}
\section{Related Work}
\vspace{-2mm}

\michael{This section should be about related work.  Explaining why we are different than PINNs or Chen should come in the intro.  If they are close enough that readers will wonder the connection, let's position ourselves with respect to both of them, and let's explain how we are different at the end of the intro.}
Physics-informed machine learning has become a rapidly growing area~\citep{karniadakis2021physics,EdwCACM22}. 
Consider, for example, \citet{raissi2019physics} and and subsequent work such as~\citet{mao2020physics,zhang2020learning,chen2020physics,penwarden2021multifidelity,lou2021physics}. 
The core idea is to use an NN to represent the solution function. 
The training objective includes a loss term to fit the boundary/initial condition and a residual term to fit the differential equation. 
The residual term is computed by applying the differential operators on the NN and then evaluating at a set of collocation points. 
The closer the residual is to zero, the more the NN surrogate fits the equation. 
While PINNs have been successfully used to solve many forward and inverse problems, the differential operators in the residual term have also brought challenges in optimization~\citep{krishnapriyan2021characterizing,wang2022and,EdwCACM22}.
This suggests the need for more refined optimization methods (such as augmented Lagrangian methods or sequential quadratic programming methods) to provide a more principled basis with which to combine domain-driven and data-driven models. 

GPs have also been used for modeling or learning from physical systems. 
Early works~\citep{graepel2003solving,raissi2017machine}  leveraged kernel differential methods to solve linear equations with observable sources: $L u =  g$, where $L$ is a linear operator, $u$ is the solution, and $g$ is the source term. 
\citet{graepel2003solving} assume the data consists of noisy observation of $g$. 
Given the covariance (kernel) function of $u$, the covariance of $g$ is obtained via kernel differentiation with operator $L$. 
Hence the kernel parameters and noise variance can be estimated from maximizing the marginal likelihood \eqref{eq:gp}.  
It is also straightforward to calculate the predictive distribution of $u$ given the observation of $g$ via cross-covariance between $u$ and $g$. 
\citet{raissi2017machine} assumed both $u$ and $g$ have noisy observations, and hence a joint GP prior over $u$ and $g$ is constructed via kernel differentiation. 
Recent work~\citep{wang2022physics} instead used a similar formulation to PINNs to conduct deep kernel learning, \ie applying differential operators on the posterior function samples.  
While it is quite effective with deep kernels, this method does not perform well when reducing the deep kernels to commonly used shallow kernels.

Recently,~\citet{chen2021solving} used kernel differentiation to solve linear and nonlinear PDEs. This work  minimizes the RKHS norm of the solution with constraints and/or regularizations that the equation is satisfied on a set of collocation points. 
From a high-level view, our method uses a similar strategy to integrate physics, \ie applying  kernel differentiation, and learning from  data fitting plus regularization on collocation points (\ie the data likelihood and virtual likelihood). 
However, both the modeling and inference are different. 
Our model is a nonparametric Bayesian model that creates a joint distribution over the solution function, its derivative functions, the noisy data and virtual observations,   while~ \citet{chen2021solving} used kernel ridge regression (square loss plus RKHS norm), a typical frequentist based kernel learning framework~\citep{kanagawa2018gaussian}.  
One might hope that that a Bayesian model is more amenable for reasoning under uncertainty. 
Second, our variational inference  estimates the posterior distribution of the solution function values and its derivatives, rather than providing a point estimation~\citep{chen2020physics}. We also find an interesting insight that the posterior process (with Gaussian variational approximations) is still a GP and maintains the kernel differentiation property. 
\cmt{
More recently,~\citet{chen2021solving} also uses kernel differentiation to calculate the covariance among the solution and its derivatives to solve linear and nonlinear PDEs. 
This work  then minimizes the RKHS norm of the solution with constraints and/or regularizations that the equation---\ie a combination of the derivatives (and solutions)---is satisfied in a set of collocation points. 
This achieves the same lever of solution accuracy as PINNs in solving several benchmark PDEs. 
Their method can also jointly estimate an unknown latent source function.  
From a high-level view, our method uses a similar strategy to integrate physics, \ie applying  kernel differentiation, and learning from  data fitting plus regularization on collocation points (\ie the data likelihood and virtual likelihood). 
However, both the modeling and inference are different. 
Our model is a nonparametric Bayesian model that creates a joint distribution over the solution function, its derivative functions, the noisy data and virtual observations,   while~ \citet{chen2021solving} used kernel ridge regression (square loss plus RKHS norm), a typical frequentist based kernel learning framework~\citep{kanagawa2018gaussian}.  
One might hope that that a Bayesian model might be more amenable for reasoning under uncertainty. 
Second, our inference aims to estimate the posterior distribution of the solution function values and its derivatives, rather than a point estimation~\citep{chen2020physics}. 
To this end, we developed a stochastic variational inference algorithm with a whitening method to estimate the posterior and the noise variance jointly from data. \citet{chen2021solving} developed a Gaussian-Newton algorithm to obtain the point estimation. 
In the experiments, we have integrated the estimated posterior to obtain the predictive distribution, which is reasonable both qualitatively and quantitatively.
See Fig. \ref{fig:pendum-exact-train}\ref{fig:pendulum-noisy-train} \ref{fig:allen-cahn} and MNLL in the tables of Sec. \ref{sec:experiments}. 
}
\michael{Nice comparison.  Can we make it shorter, but not remove substace.}

Another related work involves latent force models (LFMs)~\citep{alvarez2009latent}, which aim to integrate incomplete equations with unknown latent forces for GP learning. 
Based on the kernel of latent forces, the LFM convolves with the Green's function to derive the kernel of the target function, thereby encoding the physics into the induced kernel. 
However, LFMs are restricted to linear equations with available Green's functions. To overcome this issue, \citet{alvarez2013linear} linearized the nonlinear terms in the equation. 
\citet{hartikainen2012state,ward2020black} focused on ODEs, using a linear time-invariant (LTI) stochastic differential equation (SDE) to represent the temporal GP prior over the latent forces, and converting the original ODE  to an SDE. 
While successful, these methods do not apply to PDEs and time-spatial source functions.

 \cmt{
There are many other works proposed to estimate parameters or operators in ODEs, including \citep{calderhead2009accelerating,barber2014gaussian,macdonald2015controversy, heinonen2018learning,lorenzi2018constraining,wenk2019fast,wenk2020odin,pan2020physics}, to name a few. 
\citet{lorenzi2018constraining} proposed methods to incorporate equality or inequality constraints from ODEs and PDEs in the learning, but it demands the constraint functions have an explicit form, \ie no latent sources. 
The recent work~\citep{wang2020physics} used a similar formulation to PINNs to conduct deep kernel learning. 
It uses a hybrid Bayesian framework, where the standard GP is the conditional component, and it takes the derivative of the  random sample of the function posterior to construct a generative  component. 
The generative component resembles the residual term in the PINN loss. 
While it is quite effective with deep kernels, we found that this method does not perform well when reducing the deep kernels to commonly used shallow kernels. }
\michael{This par seems redundant with earlier material.  Can you make a good pass over this section, making it less wordy, but just as informative.}
\zhe{Sure, I will compress the related work selection altogether to fit the page limit.}


\vspace{-2mm}
\section{Empirical Results} 
\label{sec:experiments}
\vspace{-2mm}
In this section, we evaluate \ours on two illustrative and two more realistic problems.
\michael{Put in a sentence about why we chose the data we chose, and how it relates to our broad claims that we apply to everything.}
{The illustrative problems include a nonlinear pendulum system and a diffusion-reaction system, where the exact equations and the ground-truth solutions are known. 
Here, we can inspect the performance when \ours incorporates the full equation and when \ours uses only a part of the equation.
The realistic problems are the prediction tasks of metal pollution and joint motion trajectories, for which the underlying governing equations are unknown. 
Here, we examined if \ours can improve the prediction accuracy by integrating a sensible physics model (not necessarily the ground-truth equation). }

\vspace{-2mm}
\subsection{Nonlinear Pendulum} \label{sect:exp:pendulum}
\vspace{-2mm}

First, we evaluated \ours on a nonlinear pendulum system. 
Consider that a pendulum starts from an initial angle and velocity, and swings back and forth under the influence of gravity.
We are interested in how the angle $\theta$ varies with time $t$. 
The equation is given by
\begin{align}
	\frac{\d^2 \theta}{\d t^2} + \sin(\theta) 	= 0, \label{eq:ode-pen}
\end{align}
where $\sin(\theta)$ is a nonlinear term, and we choose units so that the ratio between the magnitude of gravity field and the length of the string is one. 

\begin{figure*}[t]
	\centering
	\setlength\tabcolsep{0.01pt}
	\captionsetup[subfigure]{aboveskip=0pt,belowskip=0pt}
	\begin{tabular}[c]{ccc}
		\begin{subfigure}[t]{0.33\textwidth}
			\centering
			\includegraphics[width=\textwidth]{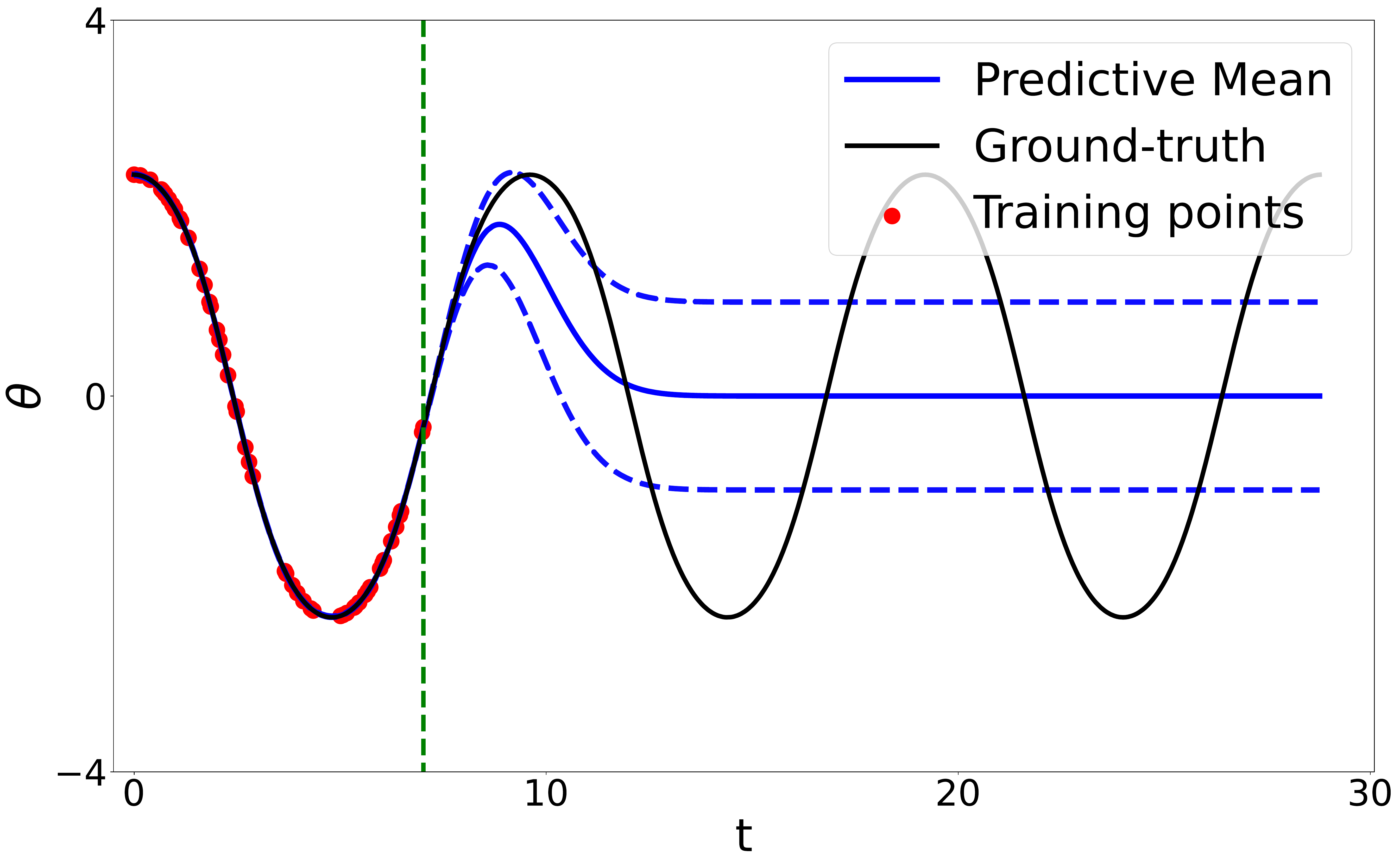}
		\end{subfigure} 
		&
		\begin{subfigure}[t]{0.33\textwidth}
			\centering
			\includegraphics[width=\textwidth]{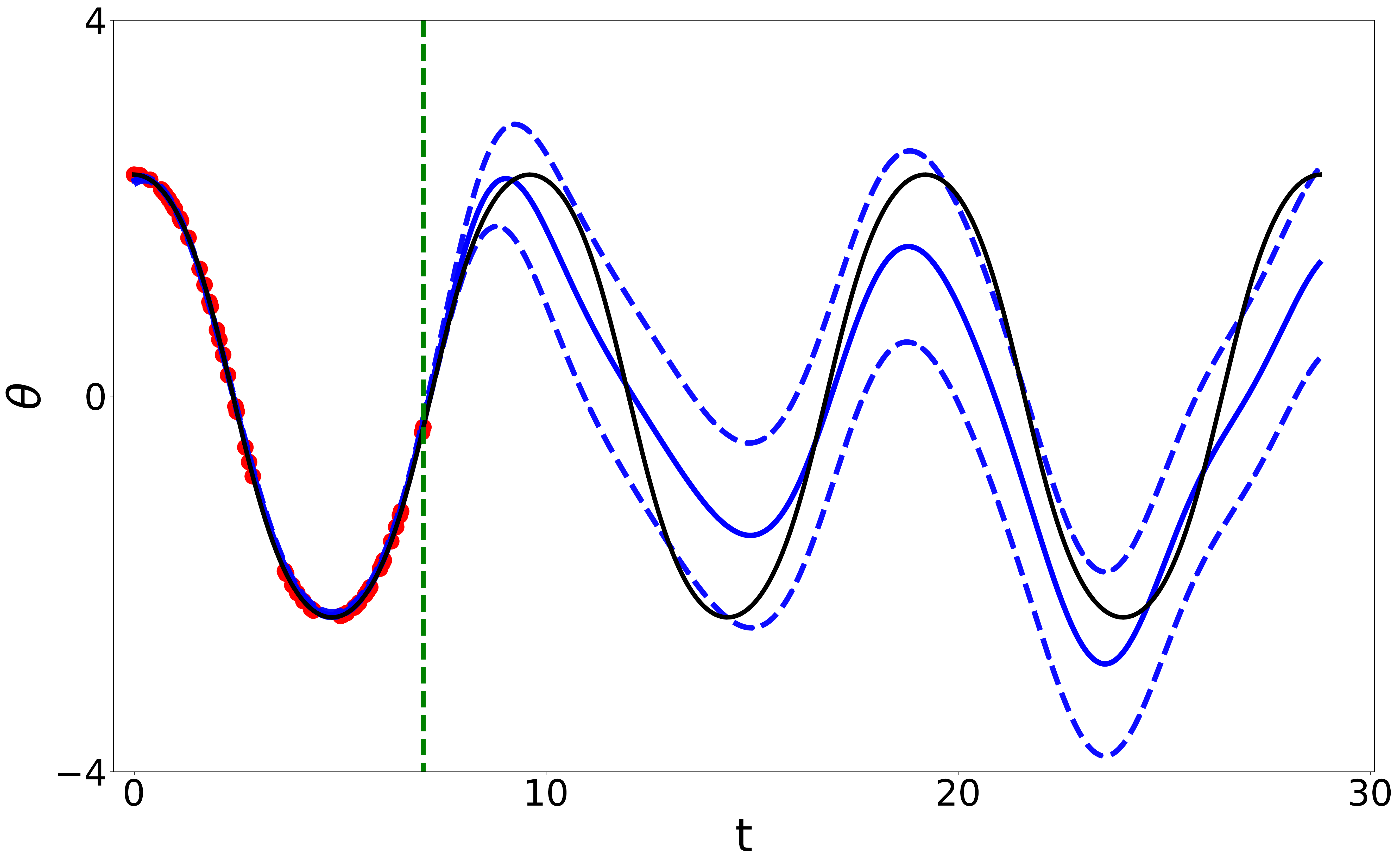}
		\end{subfigure} 
		&
		\begin{subfigure}[t]{0.33\textwidth}
			\centering
			\includegraphics[width=\textwidth]{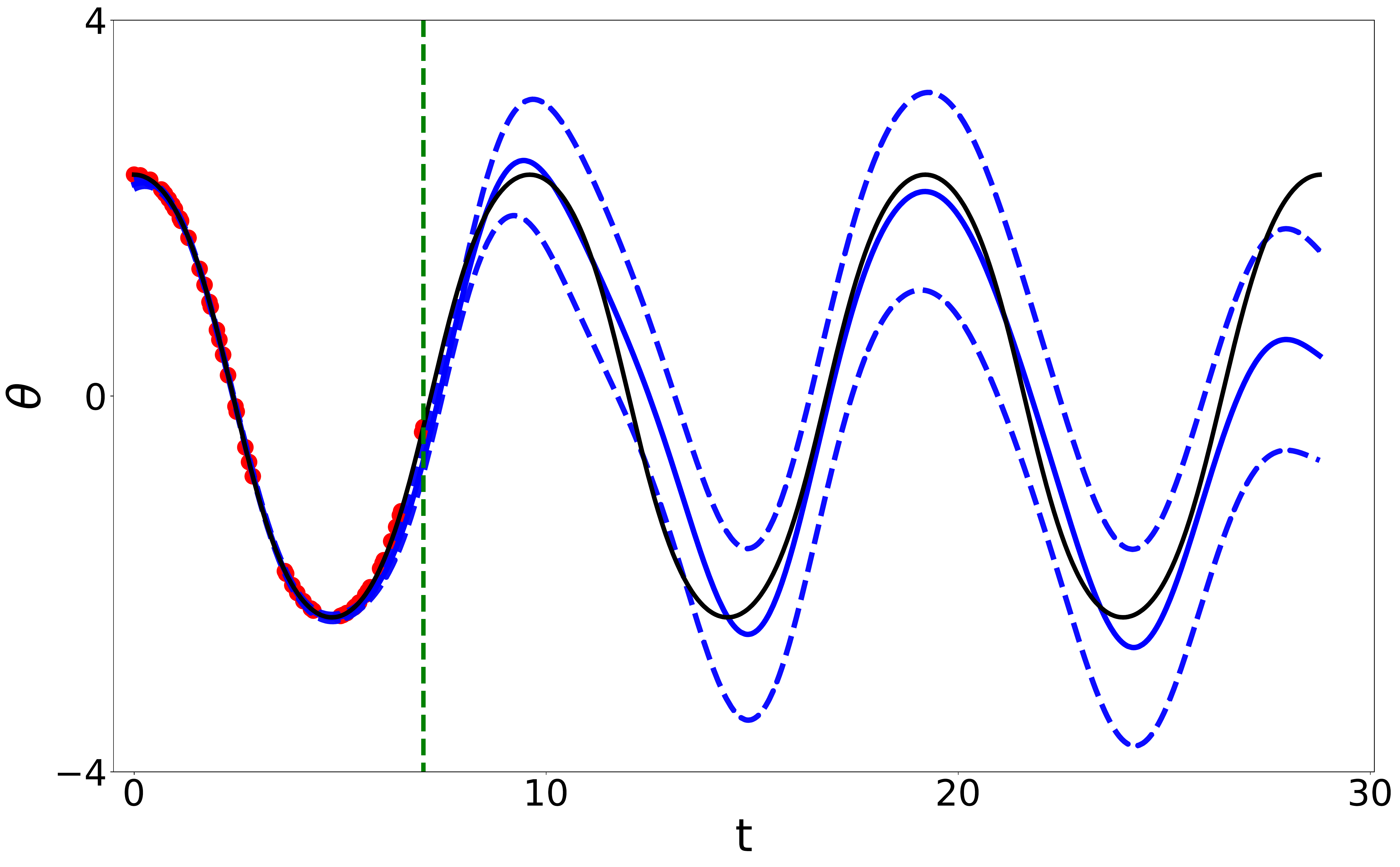}
		\end{subfigure} \\
		\begin{subfigure}[t]{0.33\textwidth}
			\centering
			\includegraphics[width=\textwidth]{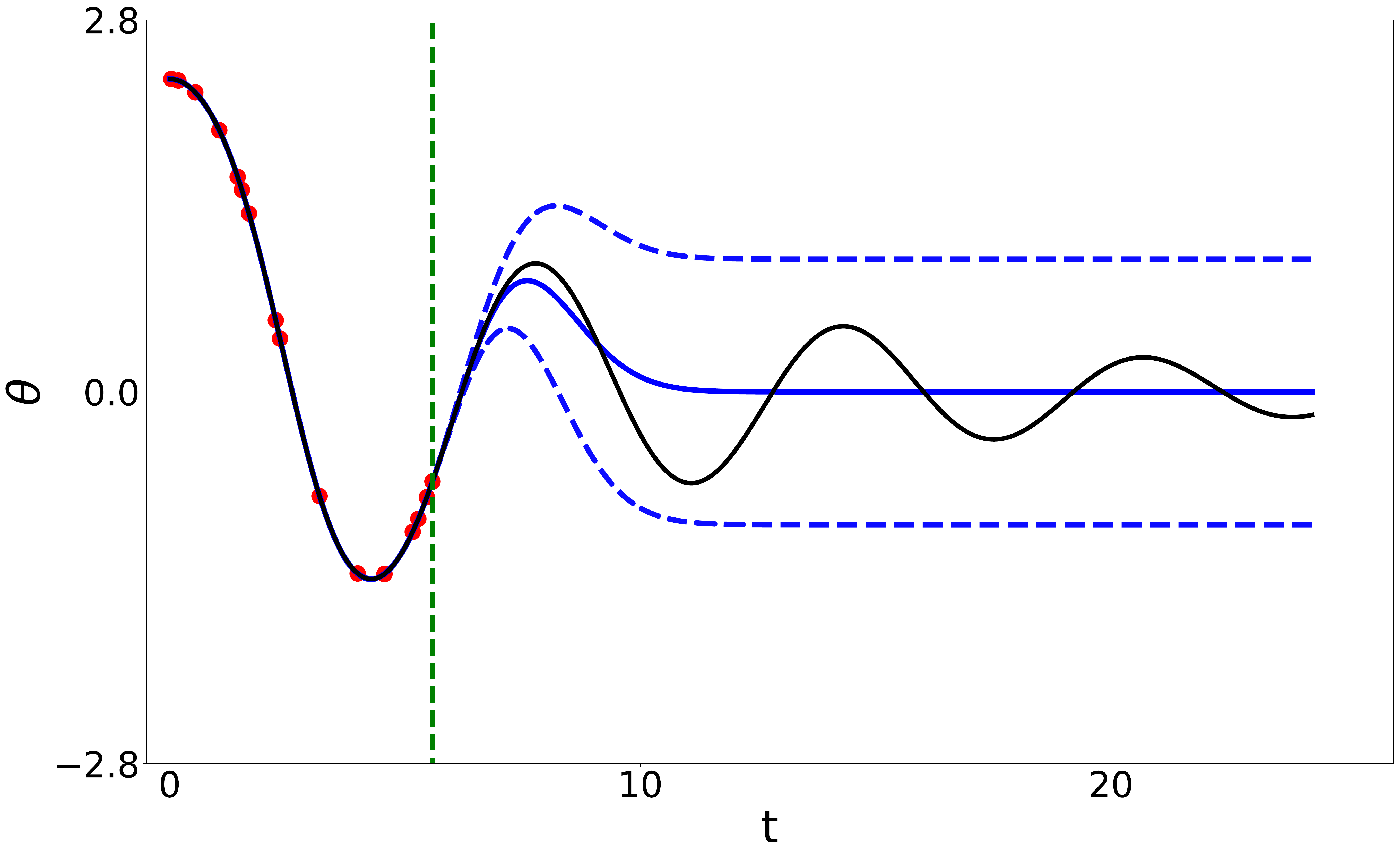}
			\caption{\small GPR} \label{fig:exact-training-gpr}
		\end{subfigure} 
		&
		\begin{subfigure}[t]{0.33\textwidth}
			\centering
			\includegraphics[width=\textwidth]{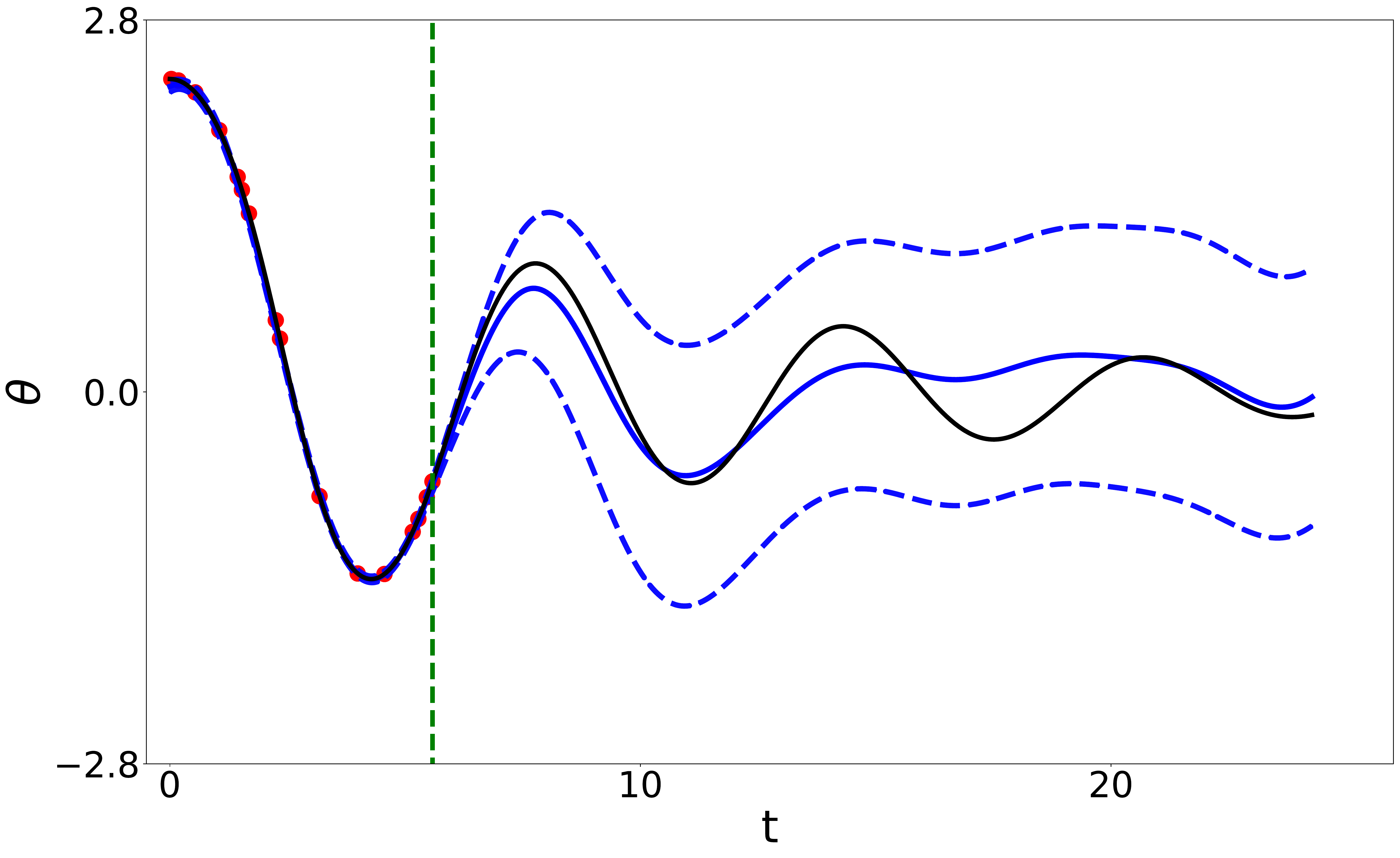}
			\caption{\small \ours-I} \label{fig:exact-training-autoip-i}
		\end{subfigure} 
		&
		\begin{subfigure}[t]{0.33\textwidth}
			\centering
			\includegraphics[width=\textwidth]{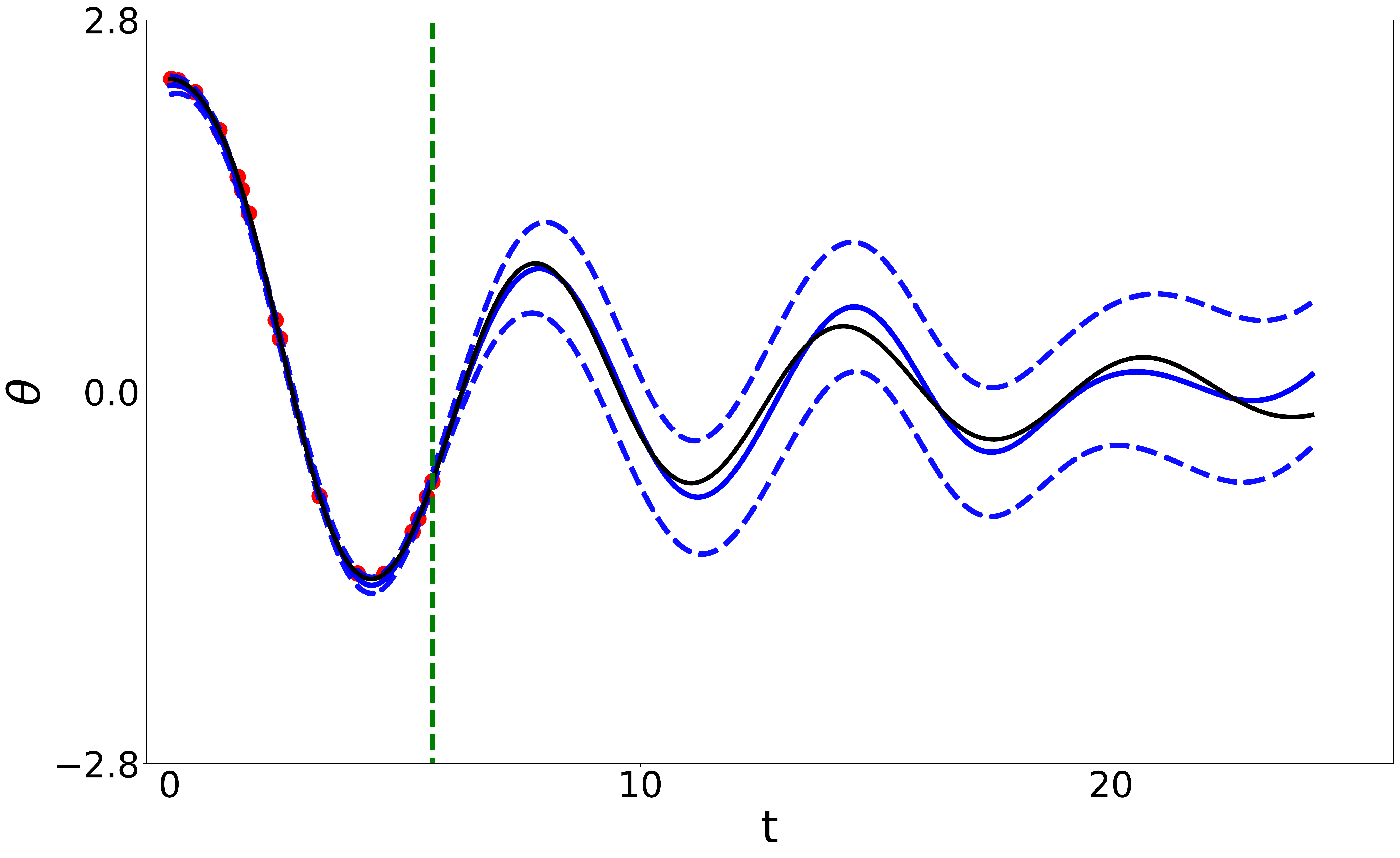}
			\caption{\small \ours-C}\label{fig:exact-training-autoip-c}
		\end{subfigure} 
	\end{tabular}
	\caption{\small Prediction in a nonlinear pendulum system with exact training examples. First row shows results without damping.  Second row shows results with damping.  Dashed lines are predictive mean $\pm$ standard deviation. Vertical line is the boundary of the training~region.} 	
	\label{fig:pendum-exact-train}
\end{figure*}

\begin{figure*}
	\centering
	\setlength\tabcolsep{0.01pt}
	\captionsetup[subfigure]{aboveskip=0pt,belowskip=0pt}
	\begin{tabular}[c]{ccc}
		\begin{subfigure}[t]{0.33\textwidth}
			\centering
			\includegraphics[width=\textwidth]{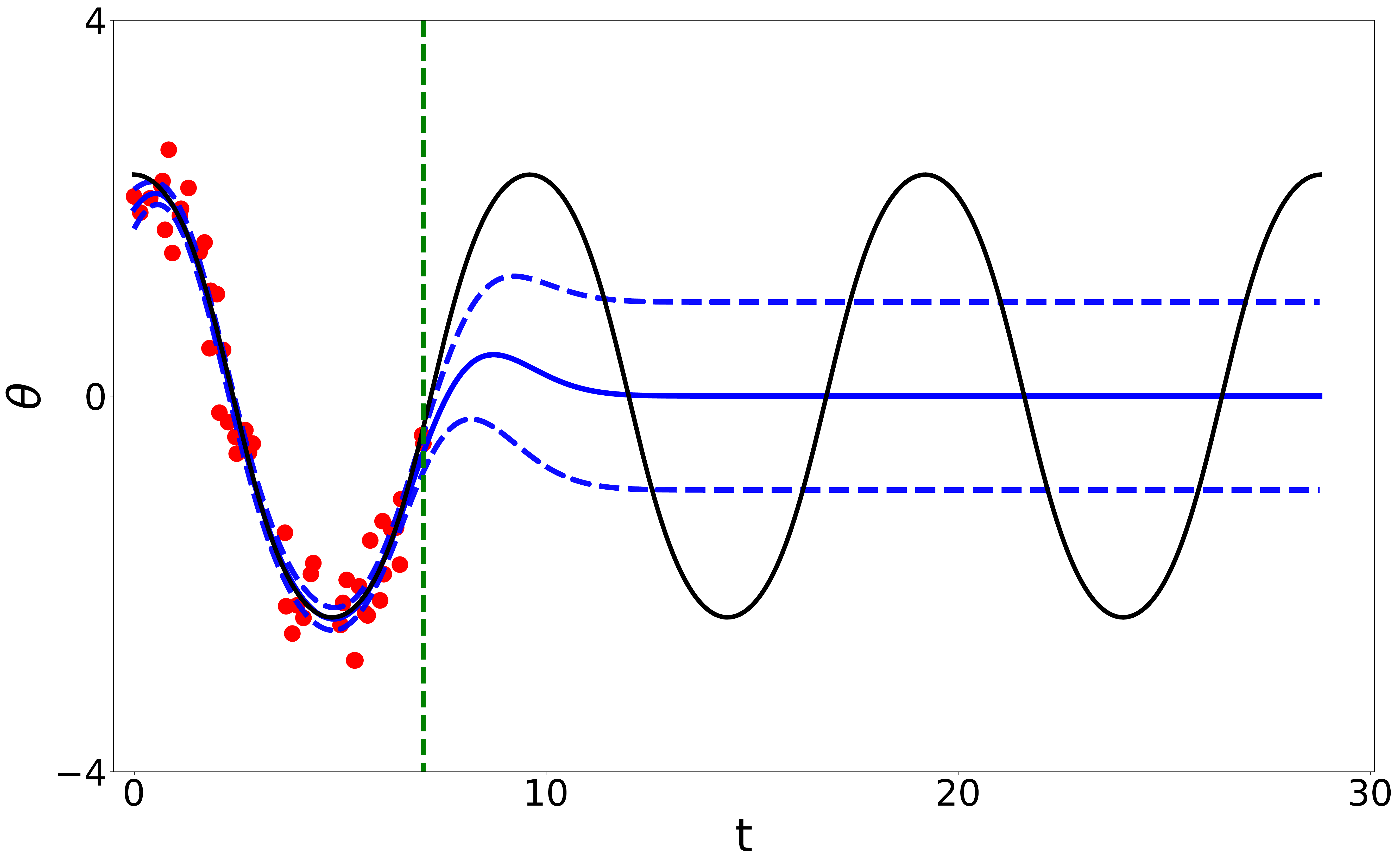}
		\end{subfigure} 
		&
		\begin{subfigure}[t]{0.33\textwidth}
			\centering
			\includegraphics[width=\textwidth]{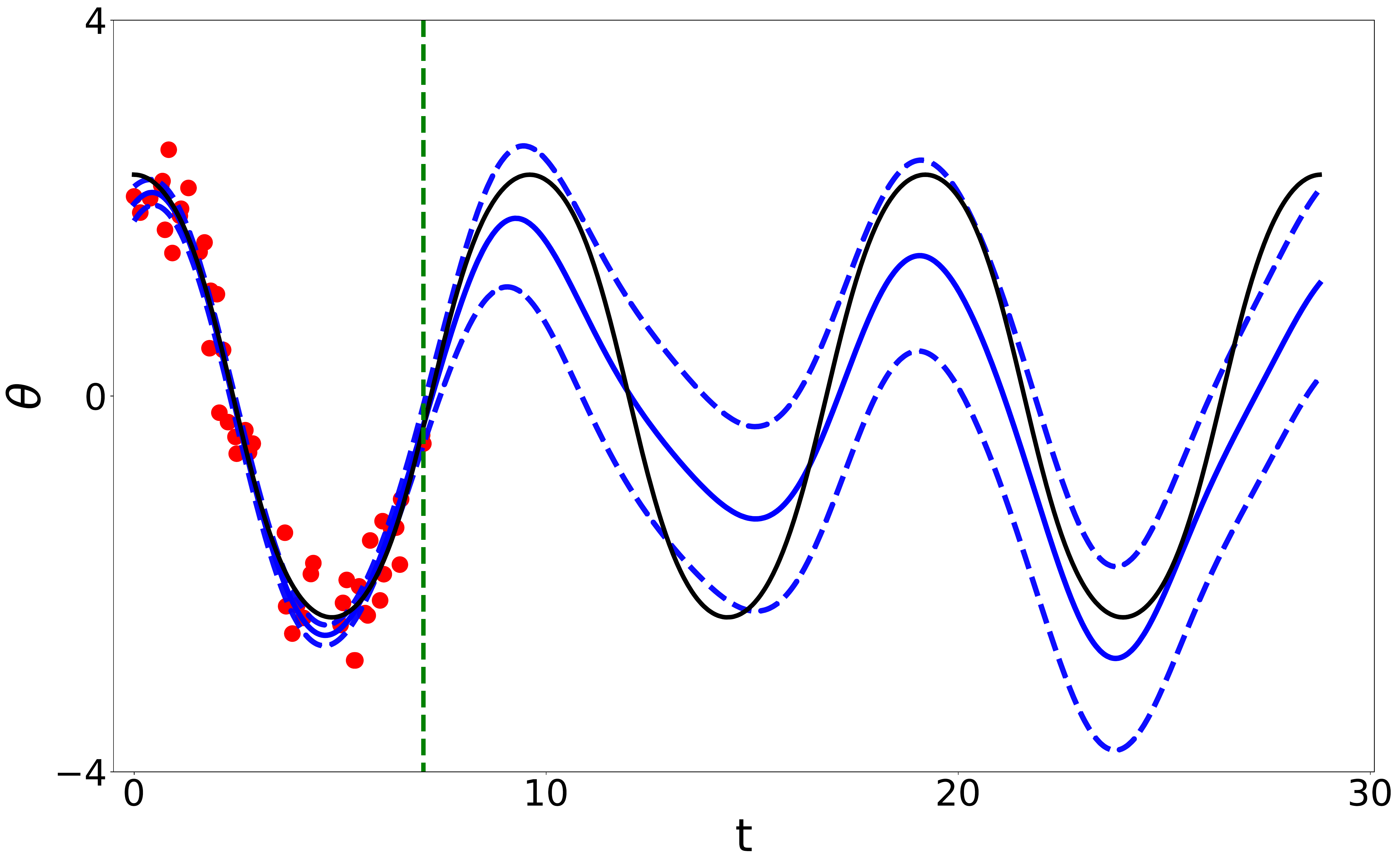}
		\end{subfigure} 
		&
		\begin{subfigure}[t]{0.33\textwidth}
			\centering
			\includegraphics[width=\textwidth]{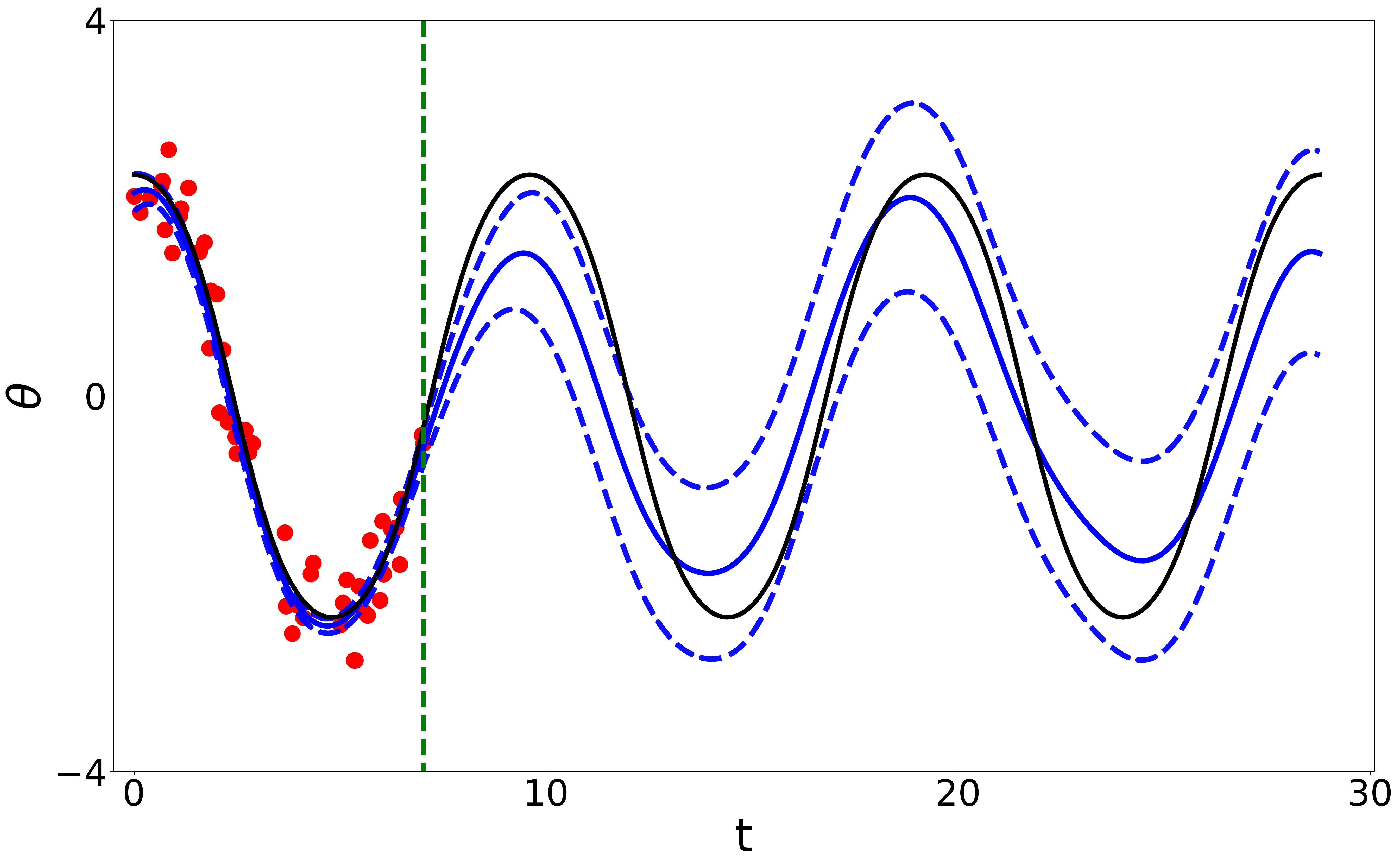}
		\end{subfigure} \\
		\begin{subfigure}[t]{0.33\textwidth}
			\centering
			\includegraphics[width=\textwidth]{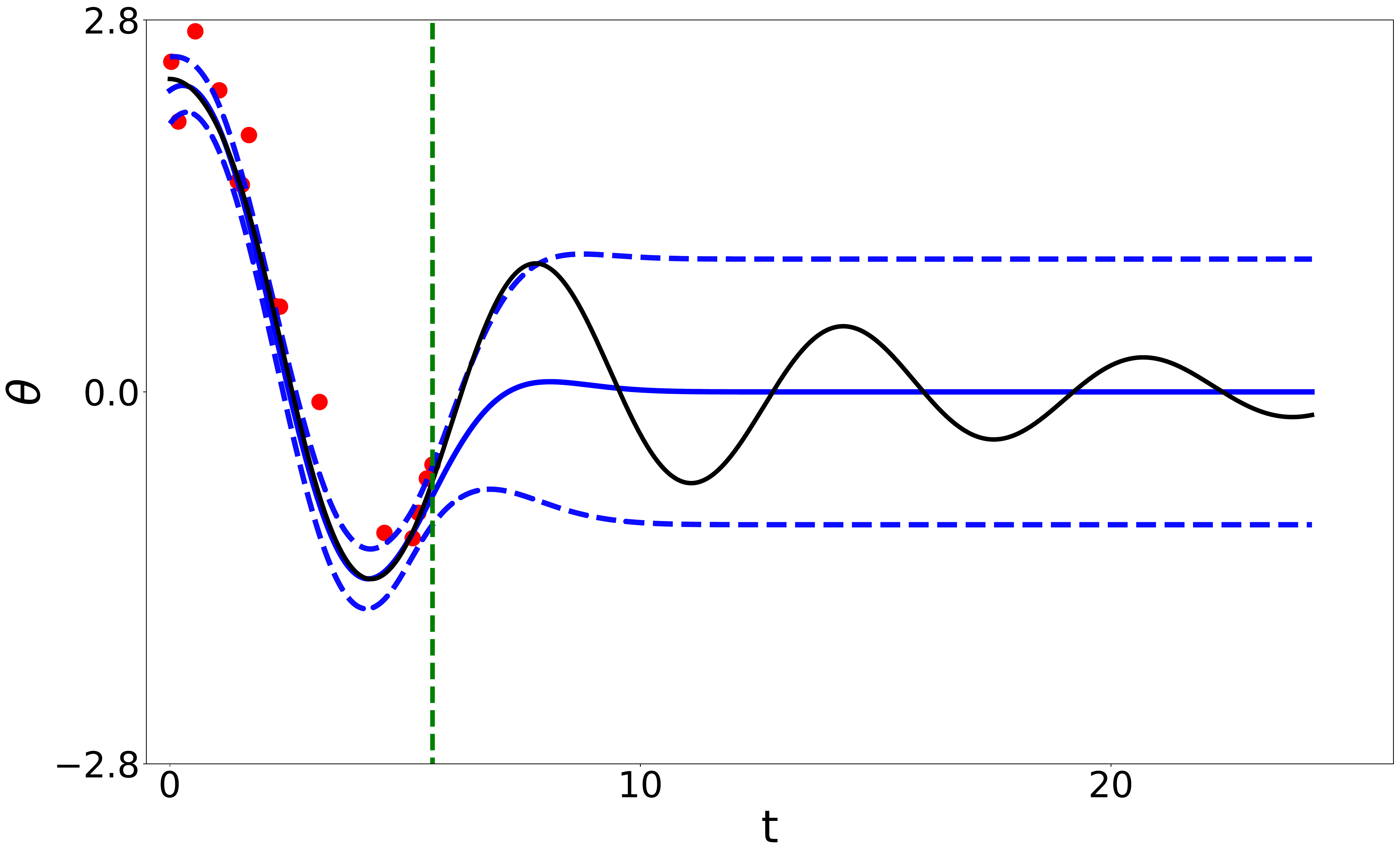}
			\caption{\small GPR} \label{fig:noisy-training-gpr}
		\end{subfigure} 
		&
		\begin{subfigure}[t]{0.33\textwidth}
			\centering
			\includegraphics[width=\textwidth]{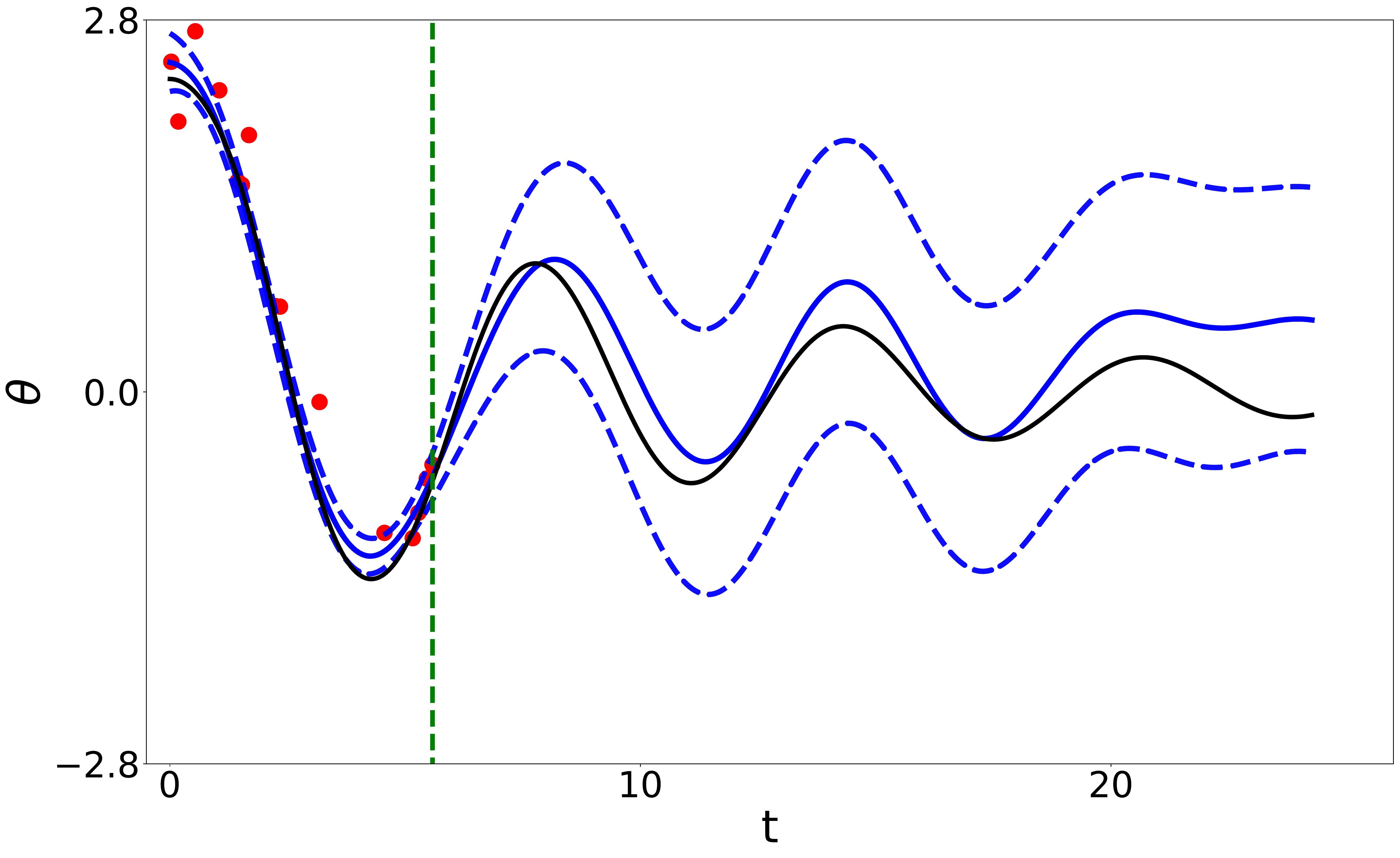}
			\caption{\small \ours-I} \label{fig:noisy-training-autoip-i}
		\end{subfigure} 
		&
		\begin{subfigure}[t]{0.33\textwidth}
			\centering
			\includegraphics[width=\textwidth]{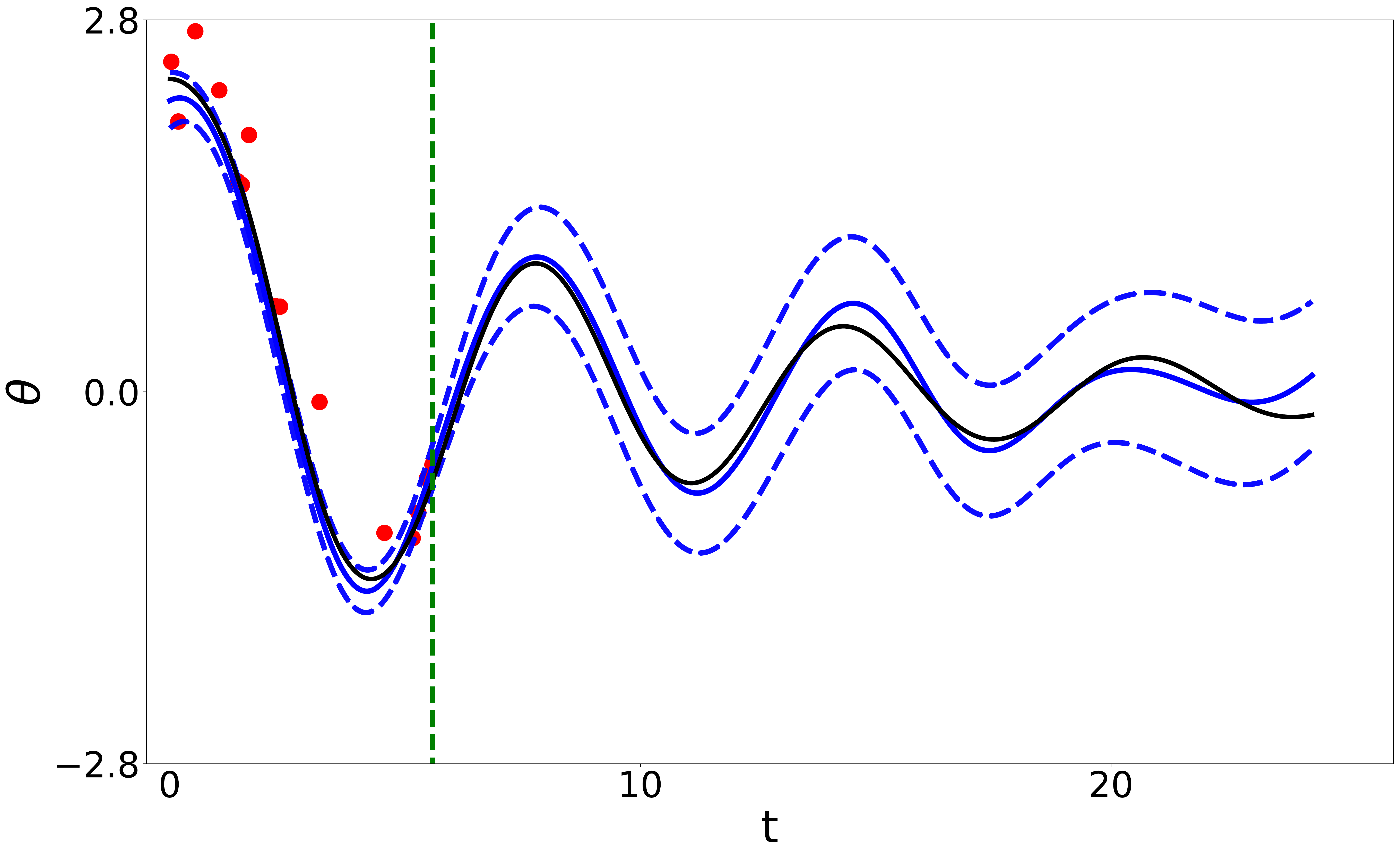}
			\caption{\small \ours-C}\label{fig:noisy-training-autoip-c}
		\end{subfigure} 
	\end{tabular}
	\caption{\small Prediction in a nonlinear pendulum system with noisy training examples. \michael{More detailed caption, that is self-contained..  Say what first and second row are, like we do in Fig 1.}First and second row show the results without damping and with damping, respectively. Dashed lines are predictive mean $\pm$ standard deviation. Vertical green line is the boundary of the training region.} 	
	\label{fig:pendulum-noisy-train}
	\vspace{-0.1in}
\end{figure*}

We set the initial angle to $\frac{3}{4}\pi$ and the initial velocity to zero. 
The change of $\theta$ exhibits apparent periodicity.
See Fig. \ref{fig:pendum-exact-train} and \ref{fig:pendulum-noisy-train} first row (the black curves). 
\michael{Make a pass over all fig references and make sure we cite consistently and clearly.  I assume this is Fig 2 and first row of Fig 3, but that is inconsistent with what I think we are using the term below.}\zhe{Yes, the references are correct; the first rows of Fig 2 and 3 are the results for incorporating equation (11); the only difference is that use uses exact training data and the other noisy training data.}
We randomly collected 50 training examples from $t \in [0, 7.3]$ that covers around $\frac{3}{4}$ period. 
Then we randomly sampled 800 test examples from $t \in [0, 28.8]$ which covers around three periods.  
We implemented both \ours and standard GPR with Pytorch~\citep{paszke2019pytorch}, and we performed stochastic optimization with ADAM~\citep{kingma2014adam}. 
We used learning rate $10^{-2}$ and ran both methods with $10$K epochs.  
To overcome the perturbation of accuracy caused by the stochastic training, we examined the prediction accuracy after each epoch and used the best accuracy for comparison.  
We used the SE-ARD kernel for both \ours and GPR, with the same initialization. 
For \ours, we let $\kappa_g$ and $\kappa_u$ share the same kernel parameters.  
We examined our method with two settings: {\ours-C}, running with the \textit{complete} equation \eqref{eq:ode-pen}; and {\ours-I}, running with an \textit{incomplete} differential equation, in which the nonlinear term is replaced by an unknown source term $g(t)$,
\begin{align}
	\frac{\d^2 \theta}{\d t^2}  + g(t) = 0. \label{eq:ode-latent-force}
\end{align}
In both settings, we randomly sampled $20$ collocation points across the whole domain $[0, 28.8]$ to integrate the equation. 
To obtain the ground-truth and to generate the training and test data, we used the \texttt{scipy} library to solve the initial value problem.  
We considered two training settings: (1) using exact training examples from the solution; and (2) using noisy training examples, where we added independent Gaussian noises sampled from $\N(0, 0.1\I)$ to the solution outputs to form the training set. 

\begin{figure*}[t]
	\centering
	\setlength\tabcolsep{0.01pt}
	\captionsetup[subfigure]{aboveskip=0pt,belowskip=0pt}
	\begin{tabular}[c]{cccc}
		\begin{subfigure}[t]{0.25\textwidth}
			\centering
			\includegraphics[width=\textwidth]{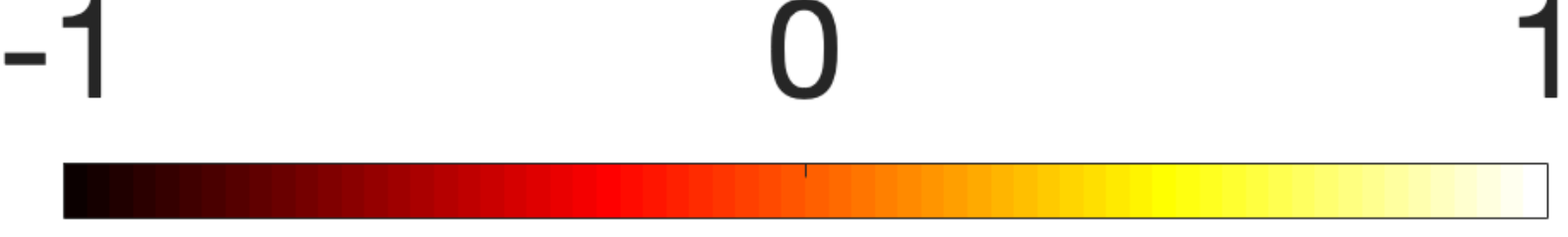}
		\end{subfigure} 
		& & &
		\\
		\begin{subfigure}[t]{0.25\textwidth}
			\centering
			\includegraphics[width=\textwidth]{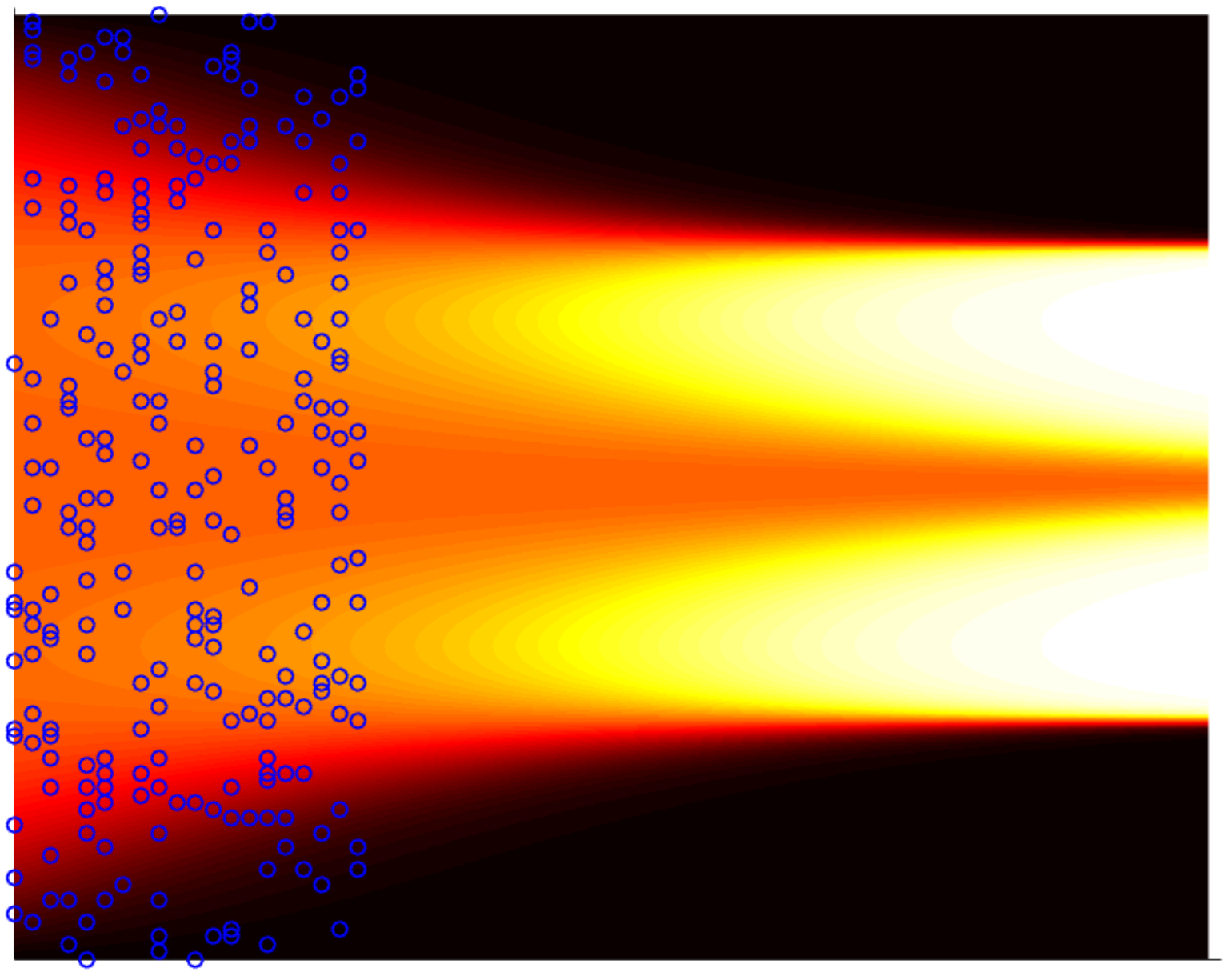}
			\caption{\small Ground-truth}
			\label{fig:allen-cahn:ground-truth}
		\end{subfigure} 
		&
		\begin{subfigure}[t]{0.25\textwidth}
			\centering
			\includegraphics[width=\textwidth]{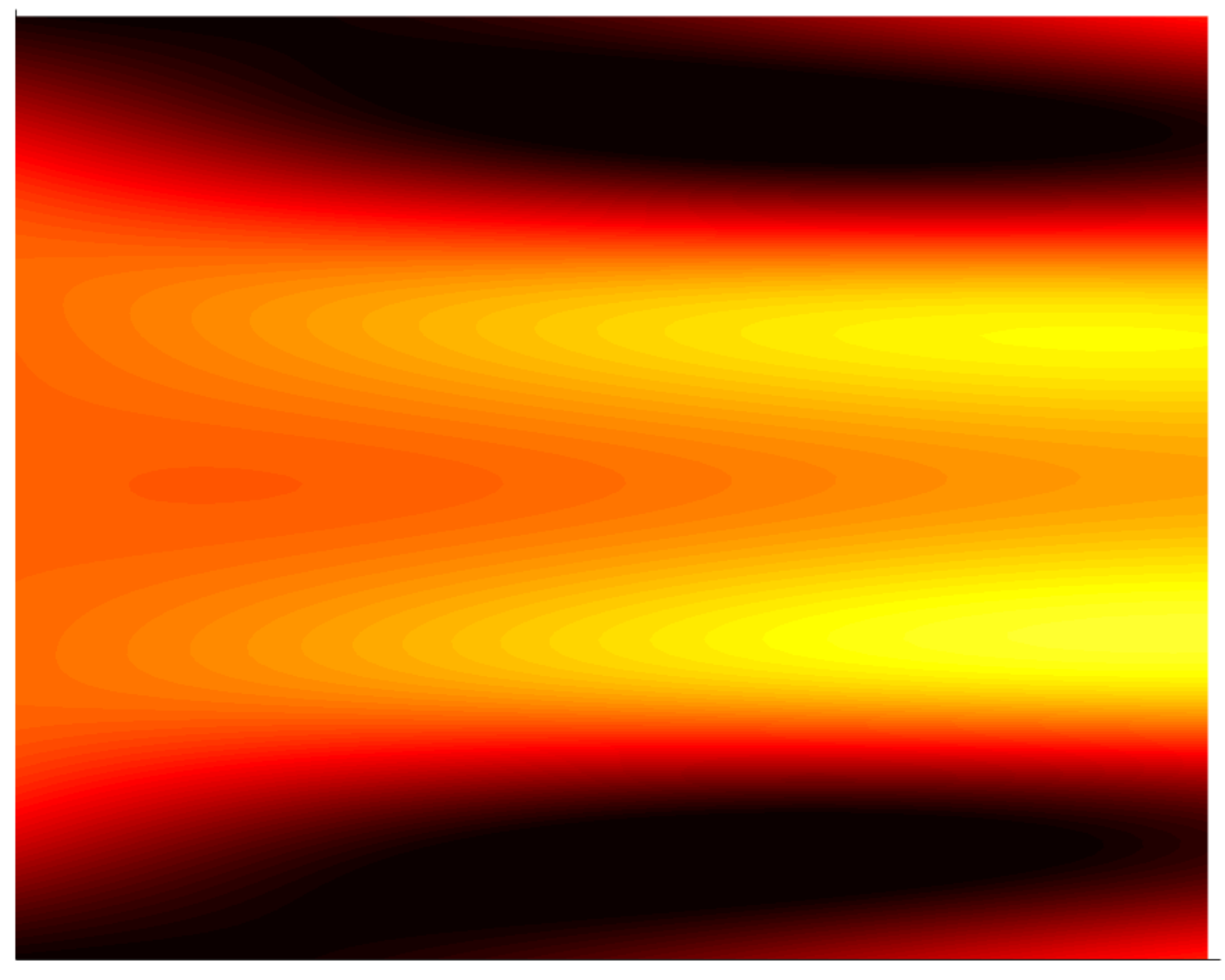}
			\caption{\small GPR: RMSE = 0.2528}
			\label{fig:allen-cahn:gpr}
		\end{subfigure} 
		&
		\begin{subfigure}[t]{0.25\textwidth}
			\centering
			\includegraphics[width=\textwidth]{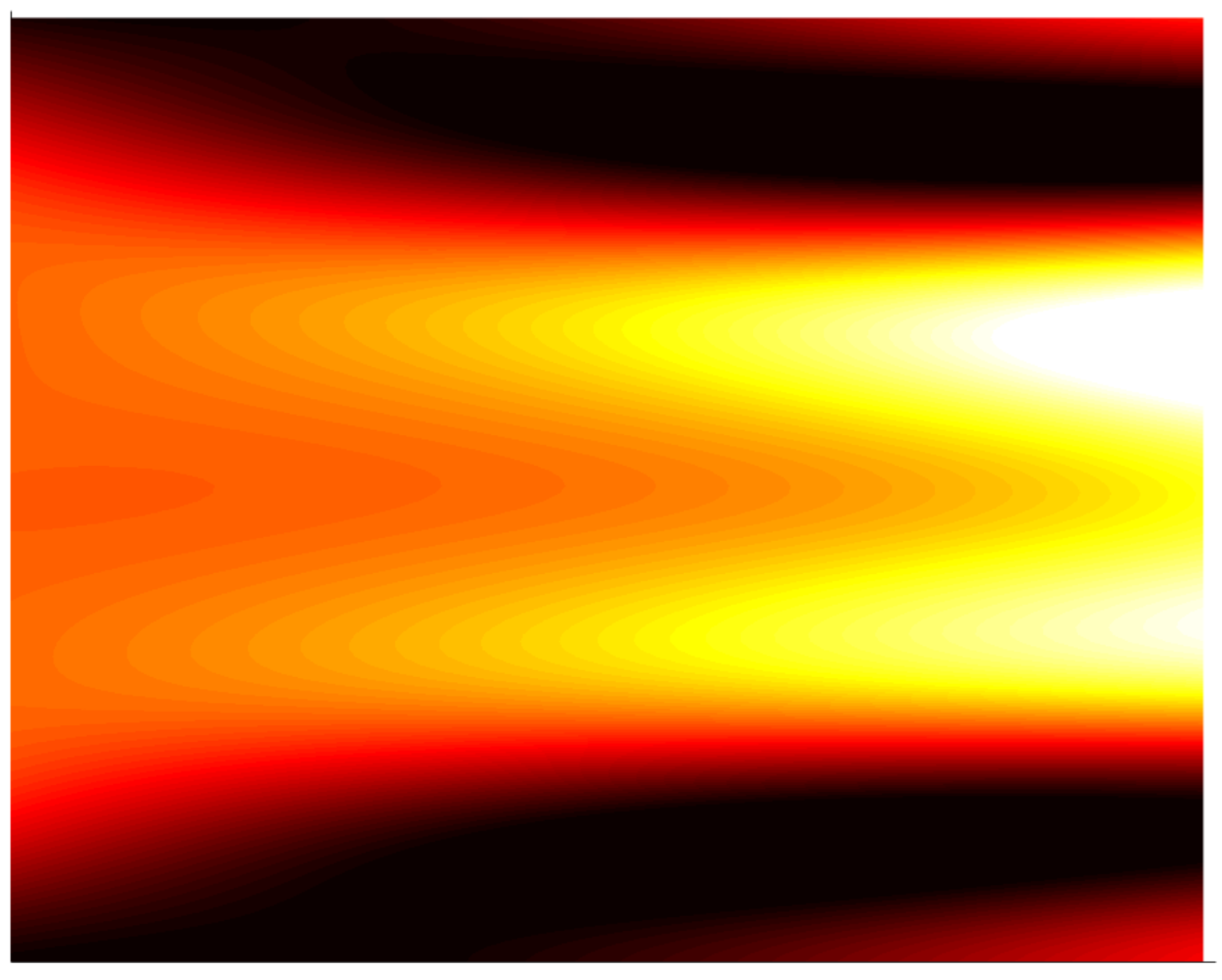}
			\caption{\small \ours-I: RMSE = 0.1869}
			\label{fig:allen-cahn:autoip-i}
		\end{subfigure} 
		&
		\begin{subfigure}[t]{0.25\textwidth}
			\centering
			\includegraphics[width=\textwidth]{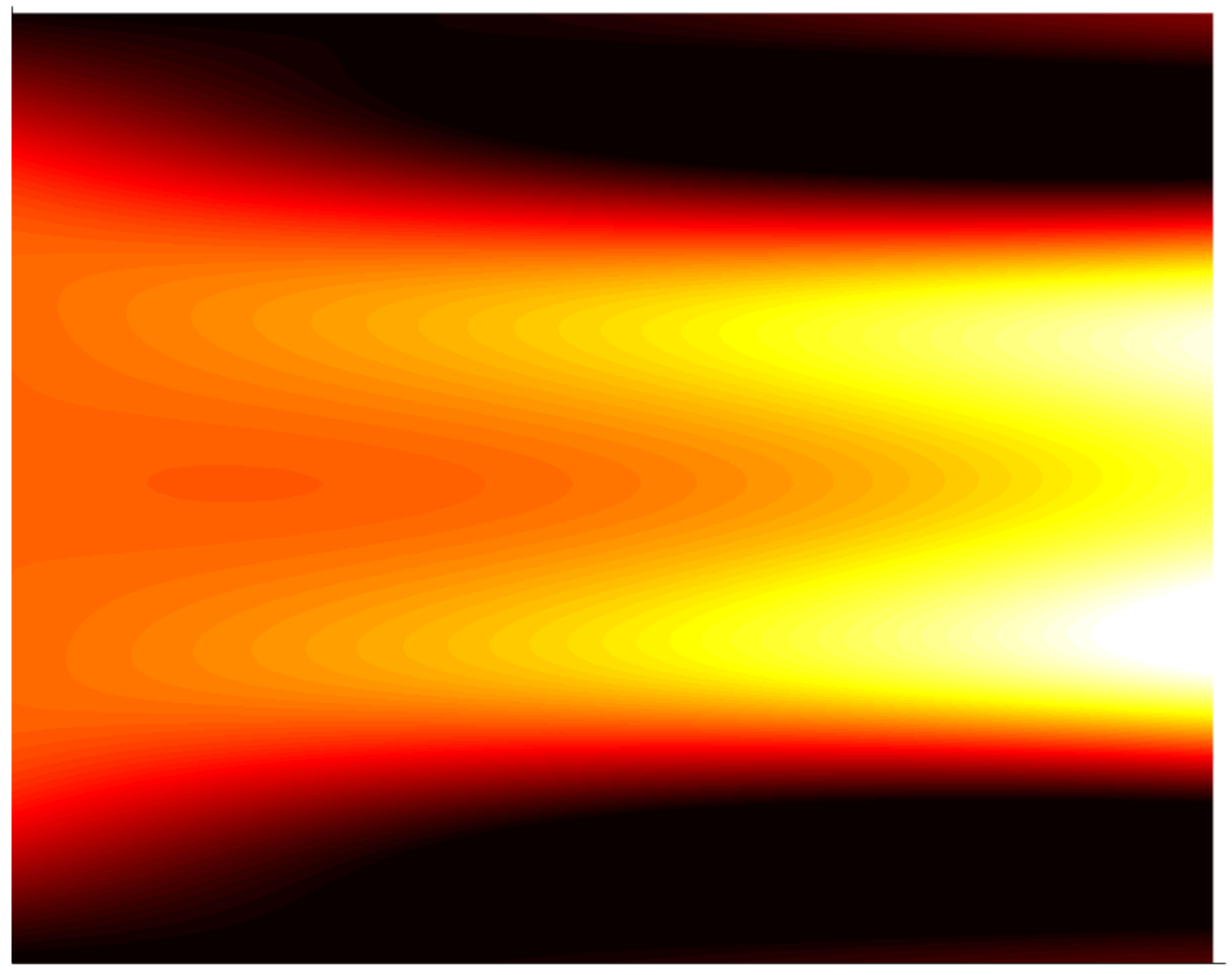}
			\caption{\small \ours-C: RMSE = 0.1865}
			\label{fig:allen-cahn:autoip-c}
		\end{subfigure} \\
		\begin{subfigure}[t]{0.25\textwidth}
			\centering
			\includegraphics[width=0.22\textwidth]{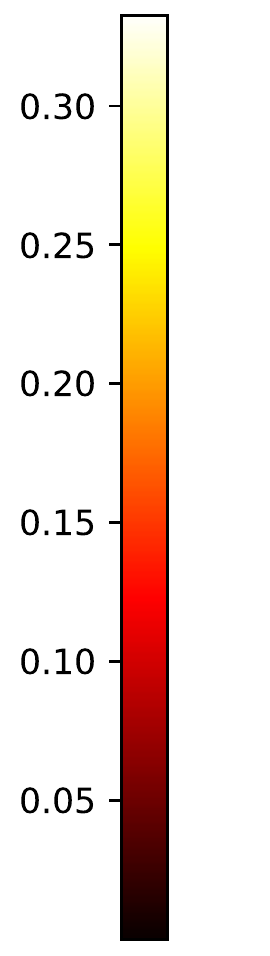}
		\end{subfigure}
		&
		\begin{subfigure}[t]{0.25\textwidth}
			\centering
			\includegraphics[width=\textwidth]{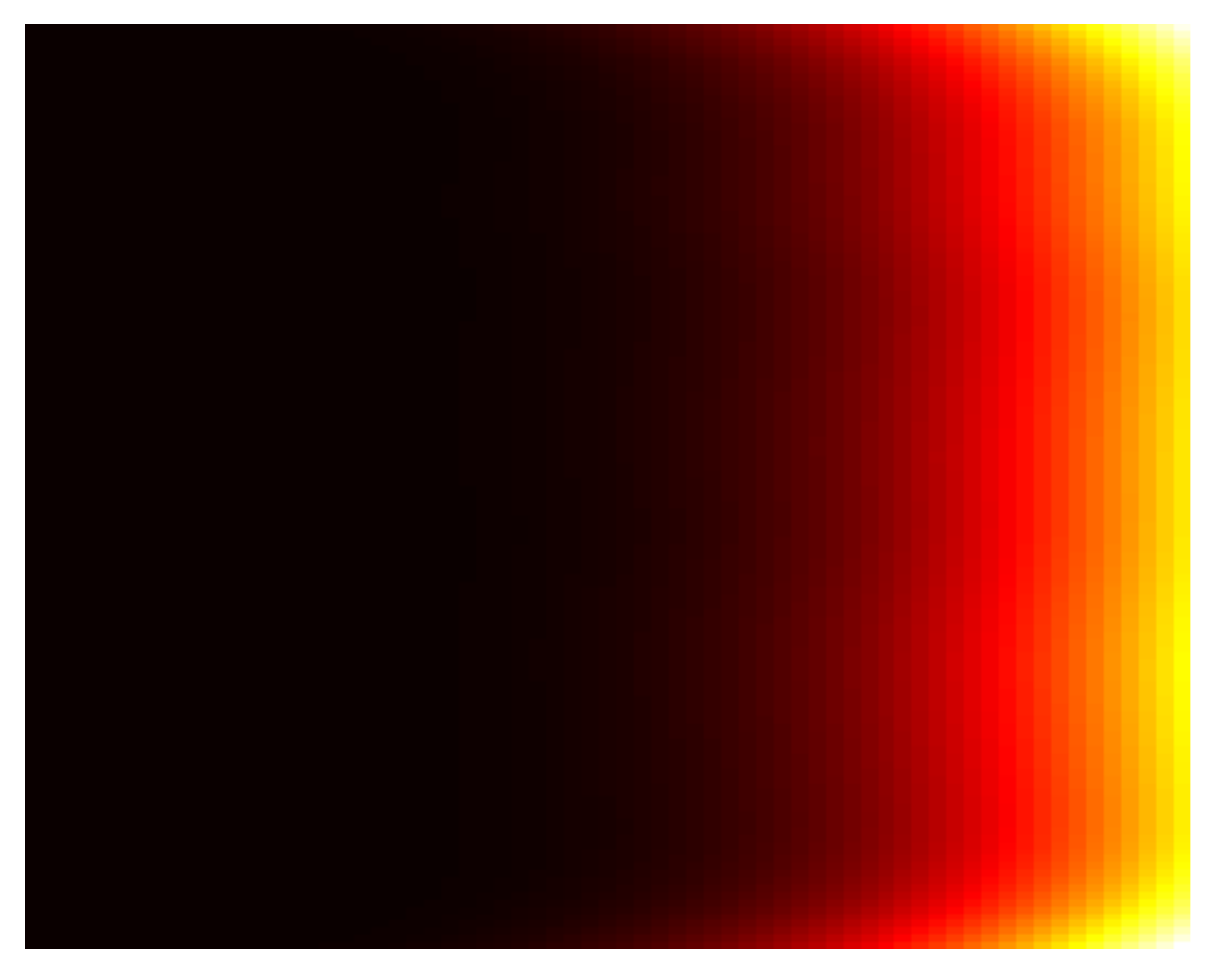}
			\caption{\small GPR predictive variance}
			\label{fig:allen-cahn:gpr-var}
		\end{subfigure} 
		&
		\begin{subfigure}[t]{0.25\textwidth}
			\centering
			\includegraphics[width=\textwidth]{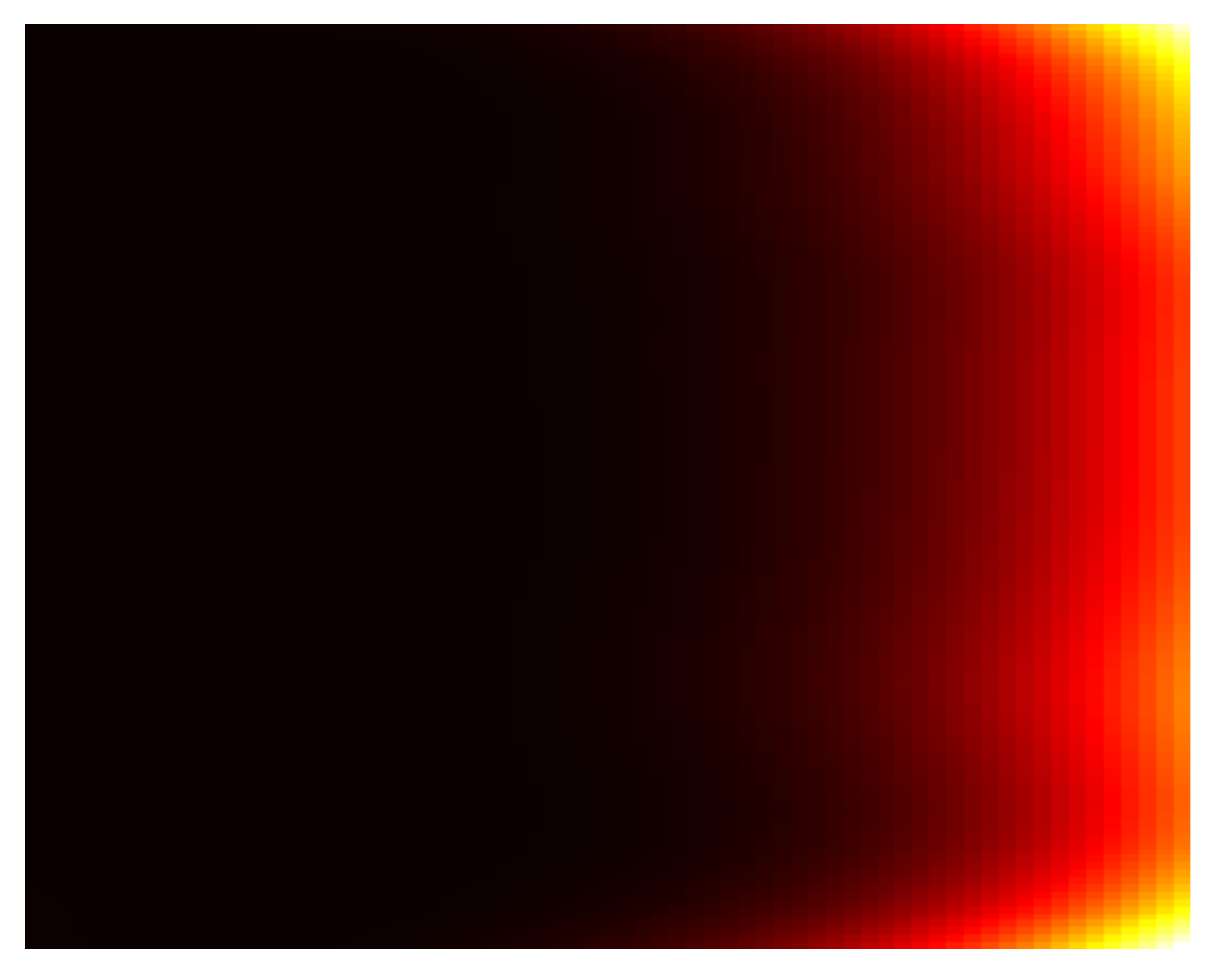}
			\caption{\small \ours-I predictive variance}
			\label{fig:allen-cahn:autoip-i-var}
		\end{subfigure} 
		&
		\begin{subfigure}[t]{0.25\textwidth}
			\centering
			\includegraphics[width=\textwidth]{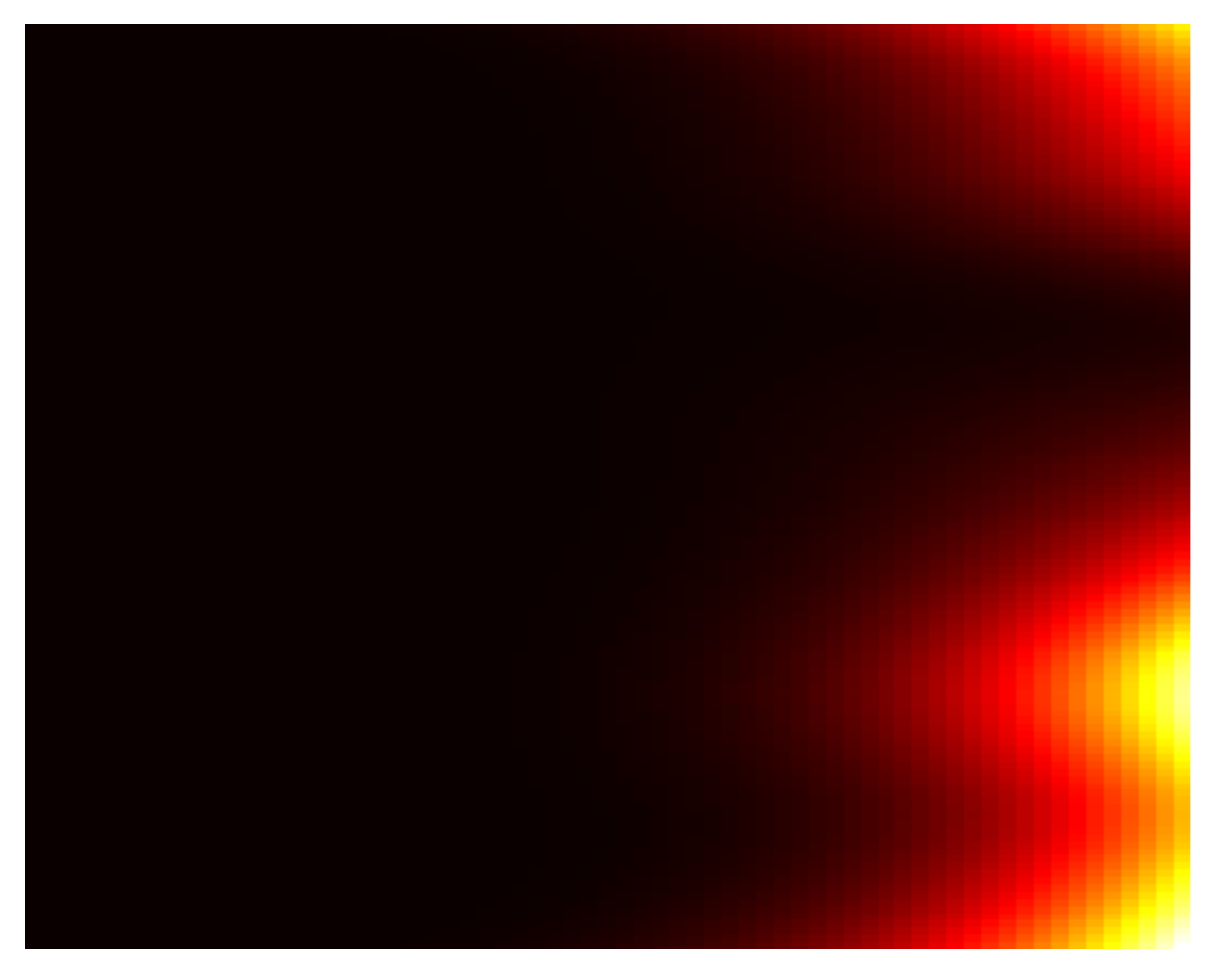}
			\caption{\small \ours-C predictive variance}
			\label{fig:allen-cahn:autoip-c-var}
		\end{subfigure} 
	\end{tabular}
	\caption{\small Prediction in a diffusion reaction system. The horizontal axis is $t$ while vertical axis is $x$. 
	\michael{Huh.  Was that cut off.  More detailed self-contained caption.}The first row consists of the ground-truth solution (where the blue circles indicate the training points used by all the methods), and the prediction made by GPR, \ours-I and \ours-C. The second row comprises of the predictive variance of each method across the domain.}
	\label{fig:allen-cahn}
\end{figure*}

In addition, we examined the case where the governing equation includes a damping term, 
\begin{align}
	\frac{\d^2 \theta}{\d t^2} + \sin(\theta) 	+ b \frac{\d \theta}{\d t}= 0, \label{eq:ode-pen-dump} 
\end{align}
where $b>0$ is a constant and was set to $0.2$. 
The damping  can be due to a type of energy loss, such as friction. 
We used the typical assumption that the friction is proportional to the  velocity. 
We randomly sampled $16$  examples from $t \in [0, 6]$ for training and $800$ examples across $t \in [0, 24.3]$ for testing. 
Again, we ran our methods in two ways. One is to integrate the complete equation \eqref{eq:ode-pen-dump}, \ie \ours-C; the other is to integrate an incomplete equation in the form of \eqref{eq:ode-latent-force}, \ie \ours-I. In the latter case, the latent source $g$ thereby subsumes both the nonlinear and damping terms. The number of collocation points was set to 20. 
\michael{That sentence is too complex and confusing, with dashes, etc.  Reword to be clear.}
While \ours-C leverages the complete equation, we do not assume the coefficient $b$ of the damping term is known. 
Instead, we view it as an unknown equation parameter, and we jointly estimate it during the inference. 
We optimized $b$ in the log domain to ensure its positiveness. 
Identical to the no-damping case,  we adopted two training settings: exact examples; and  noisy examples (with additive Gaussian noise generated from $\N(0, 0.1\I)$).  
\begin{table}[t] 
	\centering
	\small
	\begin{subtable}{0.5\textwidth}
		\centering
		\begin{tabular}{@{}lcc@{}}
			\toprule
			\textit{No damping}     &      RMSE   & MNLL    \\ \midrule
			GPR   & $1.354\pm0.005$    & $1.97\pm0.015$    \\
			\ours-I  & $0.585\pm0.017$  & $1.02	\pm0.013$        \\
			\ours-C  & $\bf{0.416\pm0.050}$  & $\bf{0.892\pm0.032}$       \\
			\midrule
			\textit{With damping}     &           \\ \midrule
			GPR   & $0.262 \pm 0.0003$     & $0.744 \pm 0.008$     \\
			\ours-I  & $0.212 \pm 0.014$  & $0.678 \pm 0.02 $         \\
			\ours-C  & $\bf{0.096 \pm 0.0035}$  & $\bf{0.155 \pm 0.01}$         \\
			\bottomrule
		\end{tabular}
		\caption{Exact training data}
	\end{subtable}
	\begin{subtable}{0.5\textwidth}
		\centering
		\begin{tabular}{@{}lcc@{}}
			\toprule
			\textit{No damping}     &      RMSE   & MNLL    \\ \midrule
			GPR   & $1.44\pm0.017$    & {$2.242\pm0.055$}    \\
			\ours-I  & $0.691\pm0.030$  & $1.206\pm0.024$        \\
			\ours-C  & $\bf{0.488\pm0.036}$  & $\bf{1.061\pm0.028}$       \\
			\midrule
			\textit{With damping}     &           \\ \midrule
			GPR   & $0.381 \pm 0.018$     & $1.07 \pm 0.029$     \\
			\ours-I  & $0.268 \pm 0.013$  & $0.937 \pm 0.011 $         \\
			\ours-C  & $\bf{0.133 \pm 0.010}$  & $\bf{0.428 \pm 0.017}$         \\
			\bottomrule
		\end{tabular}
		\caption{Noisy training data}
	\end{subtable}
	\caption{\small Prediction accuracy in nonlinear pendulum systems with/without training noise and with/without the damping term, in terms of root-mean-square-error (RMSE) and mean-negative-log-likelihood (MNLL). \ours-I and \ours-C refer to our method using incomplete and complete equations, respectively. The results were averaged over five runs.}
	\label{tb:pendulum-pred-accuracy}
	\vspace{-0.2in}
\end{table}

For each case (damping/no-damping, exact/noisy training), we repeated the experiment five times, and we examined the average Root Mean-Square-Error (RMSE), average Mean-Negative-Log-Likelihood (MNLL), and their standard deviation. 
See Table \ref{tb:pendulum-pred-accuracy}.
In all the cases, our method outperforms the standard GPR by a large margin. 
Even with an incomplete equation (including some unknown latent source), our method (\ours-I) still achieves a big improvement upon GPR, showing the advantage of effectively using physics knowledge. 
When integrating with the complete equation, our approach obtains even much better prediction accuracy (\ours-C). This is reasonable, because more precise and refined physics knowledge is leveraged. {We have also compared with physics-informed neural networks. See the results and discussion in the Appendix.}

Next, we showcase the predictive mean and standard deviation of each method of one experiment, in Fig. \ref{fig:pendum-exact-train} and \ref{fig:pendulum-noisy-train}, in contrast to the ground-truth. 
\michael{Make fig and table refs clear and consistent everywhere.}
In all cases, GPR performs well in the training region. However,  when moving away from the training region, the prediction of GPR quickly converges to the prior mean (zero), leaving a large predictive variance (uncertainty).
See Fig. \ref{fig:exact-training-gpr} and Fig. \ref{fig:noisy-training-gpr}.
\michael{Do NOT hard code subfigure references.  Use the subfigure package, and soft code them.  It's not clear whether this refers to Fig 2 and Fig 3a or to Fig 2a and Fig 3a.}By contrast, with the effective use of differential equations, \ours can predict the target function quite accurately at places very far way from the training region, exhibiting much better extrapolation performance. 
It is surprising that even with an unknown source $g$ (see  \eqref{eq:ode-latent-force}), with the key nonlinear term $\sin(\theta)$ and damping term $\theta'$ missing, \ours can still capture the variation of the target function quite well over a long range.
See Fig. \ref{fig:exact-training-autoip-i} and Fig. \ref{fig:noisy-training-autoip-i}.
\michael{Ditto, and ditto below.}
When integrating with the complete equation, \ours predicts the function values even closer to the ground-truth (see Fig. \ref{fig:exact-training-autoip-c} and \ref{fig:noisy-training-autoip-c}).  
These together have shown the advantage of \ours in effectively leveraging different equations.\michael{What is ``haven''?}\zhe{it is a typo and has been fixed.}

Finally, we examined the estimated $b$ in \eqref{eq:ode-pen-dump} by our method. 
For both noisy and exact training data, the estimation is quite good. 
For example, the estimated value by \ours-C in Fig.  \ref{fig:pendum-exact-train}c and \ref{fig:pendulum-noisy-train}c is 0.2302 and 0.2352, respectively, giving $85\%$ and $82.4\%$ relative accuracy. 
Note that we only used 16 training examples and 20 collocation points. 
The average estimation from the five experiments for exact and noisy training data is $0.228\pm 0.002$ and $0.232 \pm 0.004$, respectively.

\subsection{Diffusion-Reaction System}\label{sect:exp:allen-cahn}

We evaluated \ours on a diffusion-reaction system examined previously~\citep{raissi2019physics}.
This system is governed by an Allen-Cahn equation along with periodic boundary conditions, 
\begin{align}
	\frac{\partial u}{\partial t} - 0.0001 \frac{\partial^2 u}{\partial x^2} + 5 u^3 - 5u = 0, \label{eq:allen-cahn}
\end{align}
where $x \in [-1, 1]$, $t \in [0, 1]$, $u(0, x) = x^2 \cos(\pi x)$, $u(t, -1) = u(t, 1)$ and $u_x(t,-1) = u_x(t, 1)$.%
\footnote{We used the solution data released in \url{https://github.com/maziarraissi/PINNs}.} 
We randomly sampled 256 training examples from $t\in [0, 0.28]$, and collected 100 collocation points from the whole input domain. 
Again, we tested our method using the complete equation \eqref{eq:allen-cahn}, denoted by \ours-C, and the incomplete equation  in the form of 
\begin{align}
	\frac{\partial u}{\partial t} - 0.0001 \frac{\partial^2 u}{\partial x^2} + g(x,t) = 0, \label{eq:allen-cahn-incomplete}
\end{align}
where $g$ is an unknown source term (\ours-I). 
We ran GPR and our method for 200K epochs with the learning rate $10^{-3}$ (a larger learning rate will hurt the performance). The ground-truth solution is given by Fig. \ref{fig:allen-cahn:ground-truth}.
We show the prediction of GPR, our method with the incomplete equation (\ours-I), and our method with the complete equation (\ours-C)  in Fig. \ref{fig:allen-cahn:gpr}, \ref{fig:allen-cahn:autoip-i} and \ref{fig:allen-cahn:autoip-c}, respectively. 
As we can see, \ours is better able to capture the two reaction patterns, which look like two yellow flames. 
\michael{We should refer to fig 4a before 4b.  Probably a few sentences back, and this phrase goes there, since it is awkward here.}
GPR, however, predicts a quite uniform reaction strength, losing its time variation.  
The overall RMSE confirms that \ours achieves a much better prediction accuracy.  
In Fig. \ref{fig:allen-cahn}e-g, we show the predictive variance of each method across the domain. 
Both \ours-I and \ours-P reduce the predictive uncertainty at places distant from the training region; see the red and yellow part on the right. 
The reduction from \ours with complete equation is even more significant (\ours-C), especially at the upper-half of the right end. 


\vspace{-2mm}
\subsection{Motion Capture}
\vspace{-2mm}

We evaluated \ours on the prediction of the joint trajectories in motion capture. 
We used the CMU motion capture database.\footnote{\url{http://mocap.cs.cmu.edu/}} 
We used the trajectories of subject 35 during walking and jogging, which lasted for 2,644 seconds. We considered joint 1 and joint 50. 
From each joint, we randomly sampled 100 examples from the first half of the trajectory for training, and we randomly collected another 800 examples across the whole trajectory for testing.  
Since the ground-truth differential equation that can characterize the motions is actually unknown, we used an incomplete one with a latent source term~\citep{alvarez2009latent,alvarez2013linear},
\begin{align}
	\frac{\partial u}{\partial t} + b \cdot u(t) - c = g(t) ,
\end{align}
where $b, c>0$ are unknown coefficients and $g(t)$ is the latent source. 
In our method, we jointly estimate $b$ and $c$ in the log domain. 
To examine how the location of the collocation points will influence the performance of our method, we tested three settings: (1) \ours-T that uses the training inputs as the collocation points; (2) \ours-H that employs 200 random collocation points in the training region only, \ie half of the time span; and (3) \ours-W that employs 200 random collocation points across the whole time span of the trajectory. 
In addition to GPR, we also compared with latent force models (LFMs) proposed in~\citet{alvarez2009latent,alvarez2013linear}. 
LFMs use the kernel for the latent source $g$ and the Green's function of the equation to perform convolution so as to derive an induced kernel for $u$, which includes $b$ and $c$ as the kernel parameters. 
We also used ADAM to train LFMs. 
We ran every method for 3K epochs with learning rate $10^{-2}$, and we compared their best prediction accuracy (after each epoch). 
We repeated the experiments for five times, and calculated the average RMSE and NMLL, as listed in Table \ref{tb:motion}. 
As we can see, \ours always outperforms the competing methods. 
Since LFM on Joint 50 cannot give a reasonable test log likelihood (NMLL), we marked it as N/A, although its predictive mean is quite normal.%
\footnote{We tried a variety of learning rates and initializations, but it either ended up with a non-positive definite covariance matrix (and crashed) or with very small  test log likelihoods (10 times smaller than the competing methods), indicating a failure of learning. These might be due to some numerical issue in optimization with the induced kernel.}
We can see that \ours-T and \ours-H are comparable in most cases. 
Since their collocations points are both from the time span of the first half trajectory, this shows that the randomness of the collocation points seem not have a major influence on the predictive performance.
By contrast, \ours-W achieves much better prediction accuracy than \ours-T and \ours-H. 
This implies that a wider range of the collocation points (not the number) is more critical to improve the performance, especially in extrapolation. 
\begin{table}[t] 
    \centering
	\small
	\begin{subtable}{0.5\textwidth}
		\centering
		\begin{tabular}[c]{ccc}
			\toprule
			Method & Joint 1 & Joint 50 \\
			\hline
			GPR &  $1.727 \pm 0.026$ &  $0.257 \pm 0.007$\\
			LFM & $1.671 \pm 0.016$ & $0.257 \pm 0.006$\\
			\ours-T & $1.511 \pm 0.007$ & $0.224 \pm 0.006$\\
			\ours-H & $1.489 \pm 0.03$ & $0.225 \pm 0.005$\\
			\ours-W & $\bf{1.103 \pm 0.027}$ & $\bf{0.215 \pm 0.009}$\\
			\bottomrule
		\end{tabular}
		\caption{RMSE}
	\end{subtable}\\
	\begin{subtable}{0.5\textwidth}
		\centering
		\begin{tabular}[c]{ccc}
			\toprule
			Method & Joint 1 & Joint 50 \\
			\hline
			GPR &  $1.368 \pm 0.020$ &  $3.431 \pm 0.242$\\
			LFM & $1.721 \pm 0.020$ & N/A\cmt{$65.5 \pm 4.03$} \\
			\ours-T & $1.138 \pm 0.024$ & $2.615 \pm 0.149$\\
			\ours-H & $1.208 \pm 0.081$ & $2.664 \pm 0.154$\\
			\ours-W & $\bf{1.121 \pm 0.084}$ & $\bf{2.495 \pm 0.111}$\\
			\bottomrule
		\end{tabular}
		\caption{NMLL}
	\end{subtable}
	\caption{Prediction accuracy on motion capture datasets.} \label{tb:motion}
	\vspace{-0.3in}
\end{table}
\begin{table}[t]
\centering
	\small
	\begin{subtable}{0.5\textwidth}
		\centering
		\begin{tabular}[c]{cccc}
			\toprule
			& GPR & LFM & \ours  \\
			\hline
			Task 1 &  $0.299 \pm 0.009$ &  $0.384 \pm 0.010$ & $\bf{0.284 \pm 0.011}$ \\
			Task 2 &  $0.304 \pm 0.012$ &  $0.381 \pm 0.011$ & $\bf{0.284 \pm 0.008}$ \\
			Task 3 &  $0.232 \pm 0.009$ &  $0.358 \pm 0.005$ & $\bf{0.224 \pm 0.006}$ \\
			Task 4 &  $0.261 \pm 0.005$ &  $0.296 \pm 0.005$ & $\bf{0.247 \pm 0.004}$ \\
			\bottomrule
		\end{tabular}
		\caption{RMSE}
	\end{subtable}\\
	\begin{subtable}{0.5\textwidth}
		\centering
		\begin{tabular}[c]{cccc}
			\toprule
			& GPR & LFM & \ours  \\
			\hline
			Task 1 &  $1.16 \pm 0.064$ &  $1.36 \pm 0.058$ & $\bf{1.10 \pm 0.069}$ \\
			Task 2 &  $1.274 \pm 0.093$ &  $1.471 \pm 0.157$ & $\bf{1.219 \pm 0.129}$ \\
			Task 3 &  $0.979 \pm 0.058$ &  $1.31 \pm 0.044$ & $\bf{0.849 \pm 0.067}$ \\
			Task 4 &  $1.383 \pm 0.098$ &  $1.496 \pm 0.097$ & $\bf{1.303 \pm 0.091}$ \\
			\bottomrule
		\end{tabular}
		\caption{NMLL}
	\end{subtable}
	\vspace{-0.1in}
	\caption{Prediction accuracy on Jura datasets.} \label{tb:jura}
	\vspace{-0.2in}
\end{table}

\vspace{-2mm}
\subsection{Metal Pollution in Swiss Jura} 
\vspace{-2mm}

We evaluated \ours on an application to predict the meta concentration with the Swiss Jura dataset.\footnote{\url{https://rdrr.io/cran/gstat/man/jura.html}} 
\michael{I don't know what that means, would you explain it.}\zhe{Jura is a canton in the northwest of Switzerland.}
The dataset includes measurements of seven metals (Zn, Ni, Cr, etc.) at 300 locations in a region of 14.5 km$^2$.  
The concentration is normally modeled by a diffusion equation, $\frac{\partial u}{\partial t} = \alpha \cdot \Delta u$, where $\Delta$ is the Laplace operator, $\Delta u = \frac{\partial^2 u}{\partial x_1^2}+\frac{\partial^2 u}{\partial x_2^2} $. 
However, the dataset does not include the time information when these concentrations were measured.  
We followed prior work~\citep{alvarez2009latent} to assume a latent time point $t_s$ and estimate the solution at $t_s$, namely $h(x_1, x_2) =u(x_1, x_2, t_s)$. 
Thereby, the equation can be rearranged as,
\[
\Delta h = g(x_1, x_2)
\]
where $g(x_1, x_2) = \frac{1}{\alpha}\frac{\partial u(x_1, x_2, t)}{\partial t}|_{t = t_s} $ is viewed as a latent source term.  
Note that LFM views  $u(x_1, x_2,0)$ as the latent source, yet uses a convolution operation to derive an induced kernel for $h$ in terms of locations, where $t_s$ is considered as a kernel parameter jointly learned from data.
We tested four tasks, namely predicting: (1) Zn with the location and Cd, Ni concentration; (2) Zn with the location and Co, Ni, Cr concentration; (3) Ni with the location and Cr concentration; and (4) Cr with the location and Co concentration. 
For each task, we randomly sampled 50 example for training and another 200 examples for testing. 
The experiments were repeated for five times, and we computed the average RMSE, average NMLL and their standard deviation. 
For our method, we used the training inputs as the collocation points. 
\cmt{We compared with GPR and LFM. }
The results are reported in Table \ref{tb:jura}. 
\ours consistently outperforms the competing approaches, again confirming the advantage of our method. 
\vspace{-3mm}
\section{Conclusion}
\vspace{-2mm}
\ours is a framework for Automatically Incorporating Physics into GPs. 
This approach samples the target functions and their derivatives in a probabilistic space and uses their relationships via a virtual likelihood defined by the differential equation. 
In the future, we will use RandNLA to extend our approach in large-scale applications.

\vspace{-2mm}
\paragraph{Acknowledgements.}
This work 
was 
supported by MURI AFOSR grant FA9550-20-1-0358, NSF IIS-1910983 and NSF CAREER Award IIS-2046295. 
ASK was supported by 
LDRD
funding under Contract
Number DE-AC02-05CH11231 at LBNL and the Alvarez Fellowship in the 
CRD 
at LBNL.
MWM would like to thank the NSF, DOE, and ONR for support of this~work.

\bibliographystyle{apalike}
\bibliography{PIGP}
\newpage
\appendix
\onecolumn
\section*{Appendix}
\vspace{-2mm}
We provide additional results for comparing with physics-informed neural networks (PINNs). Since PINNs only incorporate complete differential equations and are not Bayesian methods, we examined PINNs in the nonlinear pendulum system (Sec. \ref{sect:exp:pendulum}) and the diffusion-reaction system (Sec. \ref{sect:exp:allen-cahn}) with the complete equations given, 
and report the average RMSE and standard deviation. 

\vspace{-3mm}
\section{Nonlinear Pendulum}
\vspace{-2mm}

We tested PINNs with the same four settings in Sec. \ref{sect:exp:pendulum}, namely, the equation with/without a damping term (\eqref{eq:ode-pen} and \eqref{eq:ode-pen-dump}), combined with exact/noisy training data. We used the same training and test datasets for each run. For the NN architecture, we used \texttt{tanh} activation and two hidden layers. We varied the layer width from \{5, 10, 50, 100\}. 
For a fair comparison, we tested the PINN with the same number of collocation points as used by \ours, \ie 20 points. We also ran the PINN with 10K random collocation points. For training, we first ran 1,000 ADAM epochs with  learning rate $10^{-3}$ and then ran L-BFGS with 50K maximum iterations and 50K maximum function evaluations. This is a popular practice of training PINNs \footnote{\url{https://github.com/lululxvi/deepxde}}. The implementation is based on the code of~\citet{raissi2019physics}\footnote{\url{https://github.com/maziarraissi/PINNs}}. The average RMSE for five runs and standard deviation are reported in Table \ref{tb:rmse}.
As we can see, \ours-C largely outperforms the PINN in all the cases except when the equation includes a damping term and the training data does not include any noise. In that case, the PINN with 50 or 100 neurons per layer, and using 10K collocation points can solve the equation very accurately. However, when using the same  few number of collocation points (20), the PINN with different architectures is consistently much worse than \ours, even when \ours only incorporates incomplete equations (\ie \ours-I). These results  show that the performance of the PINN can be sensitive to the architecture design, the number of collocation points, and data quality, while AutoIP is quite promising and robust to different types of data, equations, and can work well with only a small number of collocation points.

\begin{table*}[h]
	\centering
	\small
	\begin{tabular}[c]{lcccc}
		\toprule
		Method & No damping/Exact training & No damping/Noisy training & Damping/Exact training & Damping/Noisy training\\
		\midrule
		PINN-5 (20) & $1.955 \pm  0.214$ & $1.895 \pm  0.261$& $0.310 \pm  0.019$ & $0.310 \pm  0.050$\\
		PINN-10 (20) & $2.122 \pm 0.179$ & $1.824 \pm  0.231$& $0.290 \pm  0.018$& $0.342 \pm  0.020$\\
		PINN-50 (20) & $2.238 \pm 0.541$ & $1.927 \pm  0.250$& $0.297 \pm  0.044$& $0.361 \pm  0.017$\\
		PINN-100 (20) & $2.042 \pm 0.273$ & $2.407 \pm  0.353$& $0.320 \pm  0.074$& $0.384 \pm  0.066$\\
		PINN-5 (10K) & $1.479 \pm  0.115$ & $1.783 \pm  0.297$& $0.110 \pm  0.015$&$0.248 \pm  0.037$\\
		PINN-10 (10k) & $1.852 \pm  0.320$ & $1.548 \pm  0.141$& $0.049 \pm  0.023$& $0.194 \pm  0.044$\\
		PINN-50 (10k) & $1.367 \pm  0.575$ & $1.658 \pm  0.074$& ${\bf 0.00007 \pm  0.00001}$& $0.157 \pm  0.051$\\
		PINN-100 (10k) & $1.862 \pm  0.584$ & $1.993 \pm  0.357$& ${\bf 0.00007 \pm  0.00002}$& $0.186 \pm  0.045$\\
		\ours-I & $0.585 \pm 0.017$ & $0.691 \pm 0.030$& $0.212 \pm 0.014$& $0.268 \pm 0.013$\\
		\ours-C & ${\bf 0.416 \pm 0.050}$ & ${\bf 0.488 \pm 0.036}$& $0.096 \pm 0.004$& ${\bf 0.133 \pm 0.010}$\\
		\bottomrule
	\end{tabular}
	\caption{\small Root Mean Square Error (RMSE). The results were averaged over five runs. ``-\{5, 10, 50, 100\}'' mean 5, 10, 50, and 100 neurons per layer; ``(20)'' means using 20 random collocation points while ``(10K)'' means 10K collocation points. Both AutoIP-I and AutoIP-C used 20 collocation points.}
	\label{tb:rmse}
\end{table*}
\begin{table}[h]
\centering
	\small
		\centering
		\begin{tabular}[c]{ccccc}
			\toprule
			 GPR & \ours-I & \ours-C & PINN (100)  & PINN (10K)\\
			\midrule
			0.2528 &  0.1869 & 0.1865 & 0.4388 & {\bf 0.0169} \\
			\bottomrule
		\end{tabular}
	\caption{\small RMSE in the diffusion-reaction system. ``(100)'' means using 100 random collocation points while ``(10K)'' means 10K collocation points. Both AutoIP-I and AutoIP-C used 100 collocation points.} \label{tb:rmse-allen-cahn}
\end{table}

\vspace{-3mm}
\section{Diffusion-Reaction System}
\vspace{-2mm}

We used the same training and test datasets in Sec. \ref{sect:exp:allen-cahn}.  We followed \citep{raissi2019physics} to use four hidden layers with 200 neurons per layer, and \texttt{tanh} activation, to solve the Allen-Cahn equation. We tested the PINN with the same set of 100 collocation points as used by \ours, and 10K random collocation points sampled from the same domain. The training was done by first running 1,000 ADAM epochs with  learning rate $10^{-3}$ and then L-BFGS with 50K maximum iterations and 50K maximum function evaluations. 
The RMSE is given in Table \ref{tb:rmse-allen-cahn}. We can see that, with the same 100 collocation points, PINN is much worse than \ours. But with 100 times more collocation points, the PINN's performance is greatly improved. The results confirm the advantage of \ours when using a small number of  of collocation points.

\end{document}